\documentclass[conference]{IEEEtran}
\IEEEoverridecommandlockouts
\usepackage{cite}
\usepackage{amsmath,amssymb,amsfonts} 
\usepackage{algorithm}
\usepackage{algpseudocode}
\usepackage{graphicx} 
\usepackage{textcomp} 
\usepackage{microtype}
\usepackage{subfig}
\usepackage{booktabs} 
\usepackage{multirow} 
\usepackage{mdframed}
\usepackage{tabularx}
\usepackage{float}
\usepackage{placeins}
\usepackage{flushend}
\usepackage{hyperref} 
\hypersetup{hidelinks}
\newlength{\myindent}
\newtheorem{proposition}{\textbf{Proposition}}

\begin{document}

\title{A Universal Metric of Dataset Similarity for Cross-silo Federated Learning
\thanks{This work has been funded by the National Institutes of Health (NIH) under grant number R01LM014380.}
}
\author{
\IEEEauthorblockN{
    Ahmed Elhussein\IEEEauthorrefmark{1}\IEEEauthorrefmark{2}, 
    Gamze G{\"u}rsoy\IEEEauthorrefmark{1}\IEEEauthorrefmark{2}\IEEEauthorrefmark{3}
}
\IEEEauthorblockA{\IEEEauthorrefmark{1}\textit{Department of Biomedical Informatics, Columbia University, NY, USA}}
\IEEEauthorblockA{\IEEEauthorrefmark{2}\textit{New York Genome Center, NY, USA}}
\IEEEauthorblockA{\IEEEauthorrefmark{3}\textit{Department of Computer Science, Columbia University, NY, USA}}
\IEEEauthorblockA{
    ae2722@cumc.columbia.edu, 
    gamze.gursoy@columbia.edu
}
}

\maketitle 
\begin{abstract}
Federated Learning (FL) enables collaborative model training across institutions without sharing raw data, making it valuable for privacy-sensitive domains like healthcare. However, FL performance deteriorates significantly when client datasets are non-IID. While dataset similarity metrics could guide collaboration decisions, existing approaches have critical limitations: unbounded costs that lack interpretability across domains, requirements for direct data access that violate FL's privacy constraints, and poor sample efficiency. We propose a novel metric that extracts model representations after a single federated training round to predict whether collaboration will improve performance. Our approach formulates similarity assessment as an optimal transport problem with a hybrid cost function that captures both feature-level differences and label distribution divergence between clients. We ensure privacy through a careful composition of Secure Multiparty Computation (SMC) and Differential Privacy (DP) mechanisms. Our theoretical analysis establishes a formal connection between the proposed metric and weight divergence in federated training, explaining why early-round activations can predict long-term collaboration outcomes. Empirically, the metric remains tightly correlated with weight divergence throughout training, reinforcing the validity of our single-round probe. Extensive experiments on synthetic benchmarks and real-world medical imaging tasks demonstrate that our metric reliably identifies beneficial collaborations providing practitioners with an actionable tool for participant selection in cross-silo FL.

\end{abstract}
\section*{Software and data} 
Code for this paper can be found at  \url{https://github.com/G2Lab/otcost_fl}.
\section{Introduction}
Federated Learning (FL) is a distributed learning framework that enables collaborative model training without data-sharing. FL is valuable in sectors such as healthcare and insurance where data-sharing is often prohibited due to privacy concerns \cite{Huang2021-ae}. In these \emph{cross-silo} settings, FL increases sample sizes and improves model performance \cite{Zhao2020-kd}. The most common approach, Federated Averaging (FedAvg), aggregates local model updates to create a global model via weighted averaging. However, when data is collected from different sites, it is often not independent and identically distributed (non-IID). Such distribution shifts across client datasets cause divergent local model updates, degrading global performance and discouraging federated cooperation \cite{Hsieh2020-nz, li2022federated}. While personalized FL methods aim to tackle performance degradation due to non-IID data, they do not consistently outperform FedAvg. These methods often incorporate regularization parameters that control how much local client models can deviate from the global model, effectively managing the trade-off between personalization and federation. However, the optimal regularization strength requires careful tuning and is challenging in federated contexts.

Assessing dataset similarity is a practical concern for institutions that maintain long-standing partnerships and routinely collaborate on multiple studies. This information can guide task development, determine the appropriateness of FL, and inform the selection of algorithms for personalized approaches when necessary. However, diagnosing distribution shifts across institutional datasets in a federated context is challenging due to heterogeneous and often unknown \emph{site-specific} differences in data generation \cite{Zhang2022-tn, Hripcsak2013-fe}. This is further complicated by privacy constraints that prohibit direct data sharing for comparison, necessitating indirect methods to assess dataset characteristics and compatibility. If there are significant differences across client datasets, the resulting distribution shifts can lead to FL performance dropping below local training \cite{li2022federated}. For example, an international consortium of hospitals with sufficiently different diagnostic criteria may not benefit from training a shared diagnosis model \cite{futoma2020myth}. A challenge in such cases is choosing a suitable personalized approach which requires an understanding of dataset similarity across the hospitals. 

Recent native approaches to dataset similarity in FL fall into three categories: (i) gradient/weight-divergence methods that measure how client updates deviate from the global average \cite{palihawadana2022fedsim}, (ii) statistical divergence measures that quantify differences in label or feature distributions \cite{fama2024measuring}, and (iii) federated optimal transport methods like FedWaD and FedBary that compute Wasserstein distances using summary statistics of client datasets (\emph{e.g.,} quantiles and sample moments)\cite{rakotomamonjy2023federated, li2024fedbary}.

While these approaches address specific FL challenges, they face fundamental limitations. Gradient or weight divergence methods primarily quantify the extent to which client optimization processes diverge, often using vector-wide comparisons that provide insights into optimization dynamics but provide limited information about the underlying geometric structure of the client datasets. Statistical divergence and federated optimal transport approaches measures rely on summary statistics, missing fine-grained structure within the data. This limitation stems from privacy constraints in FL that traditionally prohibit sharing of raw data-points or detailed representations, forcing methods to operate on aggregated statistics that may obscure important distributional patterns. Both methods also produce unbounded similarity scores whose interpretation varies with dataset dimensionality and scale, making interpretation difficult — \emph{e.g.,}, the same cost can correspond to either gains or losses in FL performance, depending on the domain.

\textbf{Contributions:} We introduce the first dataset similarity metric designed for cross-silo FL that overcomes these limitations:

\textbf{(1) Bounded, interpretable metric:} We extract final layer representations (\emph{i.e.,}, activations) from the task model after a single federated training round to serve as learned feature representations of each client's dataset\footnote{Throughout this paper, we use the terms 'activations' and 'representations' interchangeably when referring to penultimate layer model outputs.}. We then formulate an optimal transport problem and introduce a novel hybrid cost function that operates directly on these representations, integrating two key aspects of inter-client dissimilarity: (1) feature-level differences in how clients represent individual data points, and (2) divergence in the class-conditional distributions formed by these representations. This approach results in a bounded metric within the range $[0,1]$ that offers consistent interpretation across diverse domains: costs $\leq0.2$ indicate beneficial collaboration, while costs $\geq0.3$ suggest potentially detrimental outcomes, regardless of specific dataset characteristics.

\textbf{(2) Privacy-preserving computation:} To enable the use of detailed representations while maintaining privacy guarantees, we develop a hybrid privacy-preserving protocol for our metric calculation. Specifically, we introduce a novel composition of Secure Multiparty Computation (SMC) to assess feature-level differences and Differential Privacy (DP) to assess divergence in class-conditional distributions, enabling similarity computation without exposing information from raw data or representations. We provide formal privacy analysis of this hybrid approach.

\textbf{(3) Theoretical foundations \& empirical validation:}
We establish formal links between our metric and weight divergence in federated training (Proposition \ref{prop:gradient_influence_classwise}) \emph{and show empirically that the two remain tightly correlated throughout training} (Figure \ref{fig:weight_divergence}), explaining why early-round activations can indicate whether FL is likely to suffer from performance degradation.

\textbf{(4) Practical efficiency and validation:} Our method requires only one federated training round and approximately 50 samples per class. Extensive experiments on synthetic, benchmark, and real-world medical imaging datasets demonstrate superior predictive performance compared to Wasserstein distance, with statistically significant improvements in FL outcome prediction and personalized FL regularization parameter selection.

\section{Related Work}
Assessing dataset similarity to encourage federated collaboration requires detecting distribution shifts between client datasets—a challenge complicated by data decentralization and strict privacy constraints in FL. This section reviews prior work, first examining general optimal transport methods for dataset comparison, and then discussing FL-specific approaches to measuring heterogeneity. We will then highlight how our contributions address the limitations of these methods.

\textbf{Dataset Comparison via optimal transport}. Alvarez-Melis and Fusi \cite{Alvarez-Melis_2020} introduced label-aware dataset comparisons, using optimal transport to capture differences in both feature and label distributions across datasets. Although their work offers valuable foundations, it relies on raw data access and yields unbounded costs that lack consistent interpretation across domains — both of which are problematic in federated settings.

\textbf{FL-Specific Heterogeneity Measures}. Several approaches have been developed to quantify client heterogeneity in FL. Gradient-based methods analyze how client model updates diverge during training \cite{palihawadana2022fedsim}, providing insights into optimization dynamics. However, as they use global operations on gradient/weight vectors to estimate divergence, they provide limited understanding of the underlying data distributions. Statistical approaches have shown more promise. Famá et al. \cite{fama2024measuring} evaluated nine statistical metrics and found they improve federated learning performance. However, these rely on summary statistics, and thus may miss important structural differences across datasets. 

\textbf{Federated optimal transport with encrypted moments}. FedWaD \cite{rakotomamonjy2023federated} computes Wasserstein distances using federated protocols, demonstrating applications in coreset construction. FedBary \cite{li2024fedbary} extends this with Wasserstein barycenters for client selection and data valuation. However, both methods inherit the interpretability challenges of unbounded optimal transport distances. For example, an optimal transport cost of 5.0 might indicate high similarity for MNIST but low similarity for higher-dimensional medical images. 

\section{Background}

We cover preliminaries related to the problem and our proposed solution. Note, for brevity, we leave privacy preliminaries related to SMC and DP to \cite{Evans2018-gs} and \cite{dwork2014dp}, respectively.

\subsection{Federated Learning}
In FL, a server and client network iteratively train a model. First, the server distributes model parameters; clients then train the model and send model updates back to the server; finally the server aggregates the updates. As clients optimize on non-IID data, their updates, $\Delta W^{Local}$, can diverge from global updates $\Delta W^{Global}$, leading to weight divergence \cite{zhao2018federated}, defined:
\begin{equation}
\label{eqn:weight_div}
        \frac{||\Delta W^{Global} - \Delta W^{Local}||}{||\Delta W^{Local}||}.
\end{equation}

\subsection{Optimal Transport}
Optimal transport quantifies the dissimilarity between probability distributions by defining a cost function and identifying the optimal transportation map that minimizes this cost \cite{villani2009optimal}. The \cite{kantorovich1960mathematical} formulation between two distributions $X$ and $Y$ is
\begin{equation} \label{eqnOTK}
OT(X, Y) := \min_{\pi \in \Pi(X, Y)} \int_{\mathcal{X} \times \mathcal{Y}} c(x,y) d\pi(x,y)
\end{equation}
where $c(x,y)$ is the cost function and $\Pi(X,Y)$ is the couplings over the product space $\mathcal{X} \times \mathcal{Y}$. When $c(x,y)$ is defined using euclidean distance, the cost is called the p-Wasserstein distance. For scenarios involving finite samples, there is a formulation that utilizes discrete samples. Here, the pairwise cost matrix between data points is an $n\times m$ matrix where $C_{i,j} = c(x_i,y_j)$. 

\subsection{Problem setup}
We aim to determine whether FL collaboration between clients $A$  and $B$, each with datasets $\mathcal{D}_A, \mathcal{D}_B$ of feature vectors $x_i \in \mathbb{R}^n$ and labels $y_i \in \mathbb{Z}$, outperforms local training. The potential benefits of FL depends on having similar data distributions that produce convergent model updates. Therefore, we aim to develop a similarity metric that quantifies this alignment. To be practical in cross-silo FL, any such metric must preserve privacy, be broadly applicable across diverse datasets, and exhibit sample efficiency.

\subsection{Dataset Similarity Metric}
We propose a novel metric to quantify dissimilarity between client datasets by using the global model after one round of federated training as a probe network. Our metric utilizes a per-class optimal transport framework to examine how differently clients represent the same classes, quantifying dissimilarity along two complementary dimensions: variations in feature representations and differences in class-level distributions (see Algorithm \ref{alg:ot-similarity}). A key insight is that this probe network (the global model after just one federated round) reveals fundamental differences in client data distributions that persist throughout training \cite{elhussein2025player}. This enables early prediction of collaboration success without expensive multi-round evaluation.

\paragraph{\textsc{Feature Cost}} Within each class, we measure how similarly the two clients represent individual samples. We compute feature-level dissimilarity using activation vectors from the penultimate layer of the global model after one federated learning round (Algorithm~\ref{alg:cost-matrices}). For each class $c$, feature costs are computed as pairwise cosine dissimilarities between normalized activation vectors from samples of class $c$ in dataset $\mathcal{D}_A$ and samples of class $c$ in dataset $\mathcal{D}_B$. Although pre-trained embeddings provide a model-free alternative for assessment, they lack the task-specific fidelity of our federated learning–derived representations.

\paragraph{\textsc{Label Cost}} We evaluate whether clients exhibit similar activation patterns when encoding each class(Algorithm~\ref{alg:cost-matrices}). For each label class $c \in \mathcal{Y}$, we estimate Gaussian parameters $\mathcal{S}_{d,c} = (\mu_{d,c}, \Sigma_{d,c})$ from the corresponding activation vectors in dataset $d$. The Hellinger distance between these class-specific distributions quantifies how differently the clients' models represent class $c$. Following \cite{Alvarez-Melis2018-ik}, we employ Gaussian approximations for label distributions, which provides an effective upper bound for the true cost across diverse label types.

\paragraph{\textsc{Total Per-Label Cost Matrix}}For each class $c$ shared between datasets $\mathcal{D}_A$ and $\mathcal{D}_B$, we integrate feature and label cost into a total cost matrix. The per-sample-pair cost for class $c$ is computed as: $C_{ij}^c = w_f \cdot C_{ij}^{\text{feat},c} + w_l \cdot C_{ij}^{\text{label},c}$ where $(i,j)$ indexes sample pairs with $x_i \in \mathcal{D}_A, x_j \in \mathcal{D}_B$. Since both cost components are bounded - cosine dissimilarity in $[0,2]$ and Hellinger distance in $[0,1]$ - we obtain a normalized total cost. We empirically find that a default ratio $w_f:w_l = 2:1$ effectively balances fine-grained feature representations against class-level distributional differences. This weighting can be adapted based on domain knowledge: increase $w_l$ when class imbalance is the primary concern, or raise $w_f$ when features vary significantly within shared classes. Our experiments demonstrate robustness across weight ratios from $4:1$ to $1:4$ (see sensitivity analysis in our online repository~\cite{ourrepo}). 

\paragraph{\textsc{Optimal Transport Calculation}}. We solve separate optimal transport problems for each class $c$ using the Sinkhorn algorithm with entropic regularization $\epsilon = 10^{-2}$. The final cost, $s_{AB}$, aggregates the per-class optimal transport costs weighted by the product of class sizes. This is then normalized by $(2w_f + w_l)$ to yield an interpretable cost $\tilde{s}_{AB} \in [0,1]$, where higher values indicate greater dissimilarity and thus reduced potential for collaborative benefit.

\paragraph{Sample size requirements}
Our empirical analysis reveals that approximately 50 samples per class are needed for reliable estimation. With fewer samples, the method remains valid but tends to overestimate dissimilarity due to unreliable covariance estimates. For datasets with severe class imbalance or missing classes, we only compute optimal transport for shared classes with sufficient number of samples.

\paragraph{Computational Complexity}
The computation scales efficiently with dataset size. For each class $c$, computing the cosine dissimilarity matrix requires $O(n_c^A n_c^B d)$ operations where $n_c^A$ and $n_c^B$ are the number of samples of class $c$ in datasets $A$ and $B$ respectively, and $d$ is the activation dimension. Label costs add only $O(|\mathcal{Y}|d^2)$ for per-class statistics computation. The Sinkhorn algorithm for class $c$ has complexity $O((n_c^A + n_c^B)^2\log\frac{n_c^A + n_c^B}{\epsilon})$ \cite{luo2023improved}, though, in practice, convergence is rapid for our bounded cost matrices. The total complexity is the sum across all shared classes.

\begin{algorithm}
\caption{Optimal Transport Dataset Similarity in FL}
\label{alg:ot-similarity}
\begin{algorithmic}[1]
\Require Clients $\mathcal{C} = \{A, B, \ldots\}$, datasets $\{\mathcal{D}_c\}_{c \in \mathcal{C}}$, global model $\theta^0_g$
\Ensure Pairwise cost matrix $\mathbf{S}$
\State $\triangleright$ \textbf{Phase 1: Single FL Round for Model Preparation}
\For{each client $c \in \mathcal{C}$} 
    \State $\theta_c^1 \leftarrow \textsc{LocalUpdate}(\theta^0_g, \mathcal{D}_c)$
\EndFor
\State $\theta_{\text{g}}^1 \leftarrow \textsc{FedAvg}(\{\theta_c^1\}_{c \in \mathcal{C}})$
\Statex
\State $\triangleright$ \textbf{Phase 2: Pairwise Similarity Computation}
\For{each client pair $(A, B)$ where $A, B \in \mathcal{C}$} 
    \State $\mathbf{S}[A,B] \leftarrow \textsc{PairwiseOTSimilarity}(\mathcal{D}_A, \mathcal{D}_B, \theta_{\text{g}}^1)$
\EndFor
\State \Return cost matrix $\mathbf{S}$
\end{algorithmic}
\end{algorithm}

\begin{algorithm}
\caption{Pairwise OT Similarity}
\label{alg:pairwise-ot}
\begin{algorithmic}[1]
\Require Datasets $\mathcal{D}_A, \mathcal{D}_B$, model $\theta$, feature weight $w_f$, label weight $w_l$
\Ensure OT cost $\tilde{s}_{AB} \in [0,1]$

\State $\triangleright$ \textbf{Step 1: Extract Neural Activations}
\State $(H_c, Y_c) \leftarrow \textsc{ExtractActivations}(\mathcal{D}_c, \theta)$ for $c \in \{A,B\}$

\State $\triangleright$ \textbf{Step 2: Compute Per-Class OT Costs}
\State $\mathcal{Y}_{\text{shared}} \leftarrow \{y : y \in Y_A \cap Y_B\}$ \Comment{Common classes}
\State $s_{\text{total}} \leftarrow 0$, $n_{\text{total}} \leftarrow 0$
\For{each class $y \in \mathcal{Y}_{\text{shared}}$}
    \State $H_A^y \leftarrow \{h_i : Y_A[i] = y\}$, $H_B^y \leftarrow \{h_j : Y_B[j] = y\}$ \Comment{Extract class $y$ samples}
    \State $C^y_{\text{feat}} \leftarrow \textsc{ComputeFeatureCost}(H_A^y, H_B^y)$ \Comment{Pairwise cosine costs}
    \State $C^y_{\text{label}} \leftarrow \textsc{ComputeClassCost}(H_A^y, H_B^y)$ \Comment{Hellinger distance}
    \State $C^y_{\text{total}} \leftarrow w_f \cdot C^y_{\text{feat}} + w_l \cdot C^y_{\text{label}}$ \Comment{Weighted cost matrix}
    \State $\pi^{*,y}, s^y_{AB} \leftarrow \textsc{Sinkhorn}(C^y_{\text{total}})$ \Comment{Solve OT for class $y$}
    \State $s_{\text{total}} \leftarrow s_{\text{total}} + s^y_{AB} \cdot |H_A^y| \cdot |H_B^y|$ \Comment{Weight by count}
    \State $n_{\text{total}} \leftarrow n_{\text{total}} + |H_A^y| \cdot |H_B^y|$
\EndFor

\State $\triangleright$ \textbf{Step 3: Aggregate and Normalize}
\State $\tilde{s}_{AB} \leftarrow \frac{s_{\text{total}}}{n_{\text{total}} \cdot (2w_f + w_l)} \in [0,1]$
\State \Return $\tilde{s}_{AB}$
\end{algorithmic}
\end{algorithm}

\begin{algorithm}
\caption{Feature and Class-Distribution Cost Matrix Computation}
\label{alg:cost-matrices}
\algrenewcommand{\algorithmicindent}{1.4em}
\begin{algorithmic}[1]

\State $\triangleright$ \textbf{Feature-level cost (for class $y$)}
\Procedure{ComputeFeatureCost}{$H_A^y, H_B^y$}
    \For{dataset $d \in \{A,B\}$}
        \State $\tilde{H}_d^y \leftarrow \textsc{L2Normalize}(H_d^y)$
    \EndFor
    \State $\tilde{H}_{A}^y (\tilde{H}_{B}^y)^\top \leftarrow \Call{\textsc{SecureDotProduct}}{\tilde{H}_A^y, (\tilde{H}_B^y)^\top}$
    \State $C_{\text{feat}}^y \leftarrow  1 - \tilde{H}_A^y (\tilde{H}_B^y)^\top$
    \State \Return $C_{\text{feat}}^y$
\EndProcedure

\Statex
\State $\triangleright$ \textbf{Class-distribution cost (for class $y$)}
\Procedure{ComputeClassCost}{$H_A^y, H_B^y$}
    \For{$d \in \{A,B\}$}
        \State $(\mu_{d,y}, \Sigma_{d,y}) \leftarrow \Call{\textsc{ComputeStats}}{H_{d}^y}$
        \State $\mathcal{S}_{d,y} \leftarrow (\mu_{d,y}, \Sigma_{d,y})$
        \State $\mathcal{S}^{\text{noisy}}_{d,y} \leftarrow \Call{\textsc{AddDPNoiseToStats}}{\mathcal{S}_{d,y}, \epsilon_{DP}}$
    \EndFor
    \State $h_y \leftarrow \textsc{HellingerDist}(\mathcal{S}^{\text{noisy}}_{A,y}, \mathcal{S}^{\text{noisy}}_{B,y})$
    \State $C_{\text{label}}^y[i,j] \leftarrow h_y$ for all $(i,j)$ \Comment{Constant for class $y$}
    \State \Return $C_{\text{label}}^y$
\EndProcedure
\end{algorithmic}
\end{algorithm}

\section{Privacy preservation}

Data privacy is fundamental to cross-silo FL deployment. Our contribution is a metric that achieves both strong privacy/security guarantees and practical utility through a hybrid privacy framework that applies different protection mechanisms to different components of our metric. 

For computing feature-level similarities, we adapt the SMC protocol from Du et al., \cite{Du2004-pq}. This allows exact computation of pairwise cosine similarities between representations while keeping the data completely private. The SMC protocol only reveals the final similarity values — not the underlying data or intermediate computations.

For computing label distribution differences, we apply $\rho$- zero-concentrated DP (zCDP) using the algorithm by Biswas et al. \cite{biswas2020coinpress}. Here, clients add calibrated noise to their class-conditional means and covariances with the noise magnitude is controlled by the zCDP parameter $\rho$. Note $\rho$ can be converted to meet a target ($\epsilon, \delta$)-DP guarantee through the standard conversion in Bun et al., \cite{bun2016concentrated}.

The key technical challenge lies in analyzing the composition of these two different privacy mechanisms. Standard privacy composition theorems apply to homogeneous approaches (either all SMC or all DP), but our hybrid framework requires novel analysis. Specifically, we must ensure that an adversary controlling the server cannot exploit the interaction between exact pairwise similarities (from SMC) and noisy class-conditional statistics (from DP) to reconstruct or infer sensitive information from the dataset. We prove that reconstruction remains infeasible under a semi-honest adversary model when the zCDP parameter $\rho$ satisfies:
\vspace{-1mm}
\begin{equation}
\label{eqn:privacy}
\rho < \frac{6\sqrt{d}}{n}
\end{equation}
\vspace{-1mm}
where $d$ is activation dimension and $n$ is number of samples.

\textbf{Proof sketch:}
We consider attempts to reconstruct client activation matrices (\emph{e.g.,} $H_A$) via singular vectors. The adversary observes (1) noisy per-class covariance statistics related to $H_A^T H_A$ via z-CDP, and (2) the exact cross-activation cosine similarity matrix $H_A H_B^T$ via SMC.

\begin{enumerate}
    \item \textbf{Protecting Right Singular Vectors (via DP on Covariances):} Clients release DP-noised per-class activation covariance matrices, $H_{A,c}'^T H_{A,c}' = H_{A,c}^T H_{A,c} + E$, where $E$ is DP noise. The true $H_{A,c}^T H_{A,c}$ (whose eigenvectors relate to $H_A$'s right singular vectors) has a spectral gap $\delta$ (bounded by $d/3$ for $d$-dim, row-normalized activations). If Equation~\eqref{eqn:privacy} ($\rho < 6\sqrt{d}/n$) holds, the zCDP noise $||E||$ significantly exceeds $\delta$. By Wedin's theorem, this noise ensures eigenvectors of $H_{A,c}'^T H_{A,c}'$ poorly approximate true ones, obscuring $H_A$'s right singular vectors.

    \item \textbf{Protecting Left Singular Vectors (via SMC for Cross-Activations):} SMC computes $H_A H_B^T$ without revealing raw $H_A$ or $H_B$. An adversary attempting to infer $H_A$'s left singular vectors (related to eigenvectors of $H_A H_A^T$) from $H_A H_B^T$ is hindered because $H_B$ is unknown. Furthermore, if $H_A$ and $H_B$ are dissimilar, $H_A H_B^T$ provides limited information about $H_A H_A^T$'s structure.
\end{enumerate}
Thus, even with both outputs, reconstruction of $H_A$ (e.g., via SVD) remains infeasible. Full analysis is in our repository~\cite{ourrepo}.

\section{Theoretical Insights}
\label{sec:theory_insights}
Our metric quantifies client data heterogeneity by examining the geometry of activations after one FL round. We formulate optimal transport problems on a \emph{per-class basis}, where dissimilarity is assessed by transporting mass only between samples sharing the same label across client datasets. The resulting per-class optimal transport values are then aggregated. This section establishes the theoretical rationale for the components of the per-class optimal transport cost: cosine dissimilarity for final-layer activation dissimilarity within each class, and Hellinger distance for the divergence in how clients represent the \emph{same class} via their activation distributions. Further details, including full proofs, are in our repository~\cite{ourrepo}.  

\subsection{Dissimilarity of Final-Layer Activations}
\label{subsec:activation_dissimilarity}
We consider $\ell_2$-normalized final-layer activation vectors $\mathbf{z} \in \mathbb{R}^d$ (\emph{i.e.,} $||\mathbf{z}||_2=1$). For two such activations $\mathbf{z}_i, \mathbf{z}_j$ belonging to the \emph{same class} from different clients, their cosine similarity $\cos(\theta_{\mathbf{z}_i,\mathbf{z}_j}) = \mathbf{z}_i^\top \mathbf{z}_j$ reflects their alignment. The Euclidean distance on the unit sphere, $d_S(\mathbf{z}_i, \mathbf{z}_j) = ||\mathbf{z}_i - \mathbf{z}_j||_2 = \sqrt{2(1 - \cos(\theta_{\mathbf{z}_i,\mathbf{z}_j}))}$, measures their dissimilarity.

\begin{proposition}[Concentration of inner products for independent activations]
\label{prop:orthogonality}
Let $\mathbf{z}_i, \mathbf{z}_j \in \mathbb{R}^d$ be $\ell_2$-normalized final-layer activation vectors. If $\mathbf{z}_i, \mathbf{z}_j$ are modeled as independent random vectors whose directions are isotropically distributed on $S^{d-1}$, then for $t \in [0,1]$:
$$
\Pr\left(|\mathbf{z}_i^\top \mathbf{z}_j| > t\right) \leq 2 \exp(-(d-1)t^2/2).
$$
\end{proposition}
This bound follows from standard results \cite{vershynin2018high}.
\textbf{Implication:} For a given class, if client $A$'s activations $\mathbf{z}_i$ and client $B$'s activations $\mathbf{z}_j$ for that class are derived from unrelated underlying features, they are expected to be nearly orthogonal. Significant non-orthogonality suggests shared structure in representing that class. The spherical distance $d_S(\mathbf{z}_i, \mathbf{z}_j)$ quantifies this.

\subsection{Motivating Per-Class Cost Components}
\label{subsec:cost_motivation_per_class} 
The final layer typically computes $W\mathbf{z}$, followed by softmax and cross-entropy loss $\ell(W\mathbf{z}, y)$. The gradient of $\ell$ for sample $(\mathbf{z},y)$ is $\nabla_W \ell(W\mathbf{z}, y) = (\mathbf{p}(\mathbf{z};W) - \mathbf{e}_y) \mathbf{z}^\top$.

We analyze the difference in gradient contributions for two samples $(\mathbf{z}_c, y_c)$ and $(\mathbf{z}_k, y_k)$ from clients $A$ and $B$. For our per-class optimal transport, we focus on $y_c = y_k = y_{\text{target}}$.
\begin{proposition}[Gradient Dissimilarity for Same-Class Samples]
\label{prop:gradient_influence_classwise} 
The Frobenius norm of the difference between gradient contributions from $(\mathbf{z}_c, y_{\text{target}})$ (client $A$) and $(\mathbf{z}_k, y_{\text{target}})$ (client $B$), denoted $G_{ck|y_{\text{target}}} = (\mathbf{p}_A(\mathbf{z}_c) - \mathbf{e}_{y_{\text{target}}})\mathbf{z}_c^\top - (\mathbf{p}_B(\mathbf{z}_k) - \mathbf{e}_{y_{\text{target}}})\mathbf{z}_k^\top$, is bounded as:
\vspace{-1mm}
\begin{multline}
    ||G_{ck|y_{\text{target}}}||_F \leq ||\mathbf{p}_A(\mathbf{z}_c) - \mathbf{e}_{y_{\text{target}}}||_2 ||\mathbf{z}_c - \mathbf{z}_k||_2 + \\
    ||\mathbf{p}_A(\mathbf{z}_c) - \mathbf{p}_B(\mathbf{z}_k)||_2.
\end{multline} 
The term $||\mathbf{e}_{y_c} - \mathbf{e}_{y_k}||_2$ has vanished as $y_c=y_k$.
\end{proposition}

\textbf{Interpreting terms}: Proposition \ref{prop:gradient_influence_classwise} indicates that gradient differences for samples of the same class $y_{\text{target}}$ are driven by:
\begin{enumerate}
    \item \textbf{Activation Dissimilarity ($||\mathbf{z}_c - \mathbf{z}_k||_2$):} The spherical distance $d_S(\mathbf{z}_c,\mathbf{z}_k)$ between activations. This forms the feature cost component of our metric.
    \item \textbf{Prediction Difference ($||\mathbf{p}_A(\mathbf{z}_c) - \mathbf{p}_B(\mathbf{z}_k)||_2$):} This term reflects how differently the model, using client-specific information implicitly encoded in activations, predicts for these two activations of class $y_{\text{target}}$. This divergence is related to how differently clients $A$ and $B$ represent class $y_{\text{target}}$ in the activation space. If client $A$'s activation distribution for $y_{\text{target}}$ ($\mathcal{S}_{A,y_{\text{target}}}$) differs from client $B$'s ($\mathcal{S}_{B,y_{\text{target}}}$), this contributes to prediction differences. The Hellinger distance $H(\mathcal{S}_{A,y_{\text{target}}}, \mathcal{S}_{B,y_{\text{target}}})$ (per \cite{Alvarez-Melis2018-ik}) quantifies this representational divergence. While the prediction difference also depends on $W$, we use $H(\mathcal{S}_{A,y_{\text{target}}}, \mathcal{S}_{B,y_{\text{target}}})$ as a more direct, data-intrinsic measure of class representation dissimilarity for our label-aware cost component.
\end{enumerate}
This decomposition aligns with the two terms in our metric.

\subsection{Optimal Transport for Aggregating Per-Class Heterogeneity}
\label{subsec:ot_cost_per_class} 
For each class $c \in \mathcal{Y}$ present in both client datasets $\mathcal{D}_A$ and $\mathcal{D}_B$, we define an optimal transport problem. The cost of matching sample $i$ (activation $\mathbf{z}_i$) from client $A$'s portion of class $c$ with sample $j$ (activation $\mathbf{z}_j$) from client $B$'s portion of class $c$ is:
\begin{equation}
    \label{eq:per_class_ot_cost_definition}
    C_{ij}^{(c)} = w_f \cdot d_S(\mathbf{z}_i,\mathbf{z}_j) + w_l \cdot H(\mathcal{S}_{A,c},\mathcal{S}_{B,c}).
\end{equation}
Here, $d_S(\mathbf{z}_i,\mathbf{z}_j)$ is the spherical distance. $\mathcal{S}_{A,c}$ denotes the Gaussian parameters $(\mu_{A,c}, \Sigma_{A,c})$ of activations from client $A$ for class $c$ (similarly for $\mathcal{S}_{B,c}$). $H(\mathcal{S}_{A,c},\mathcal{S}_{B,c})$ is the Hellinger distance between these same-class activation distributions. Note that $H(\mathcal{S}_{A,c},\mathcal{S}_{B,c})$ is constant for all pairs $(i,j)$ within class $c$. The weights $w_f, w_l > 0$ balance feature and class-representation dissimilarities.

For each class $c$, an optimal transport plan $\pi^{(c)}$ is found that minimizes $\sum_{i,j \in \text{class }c} \pi^{(c)}_{ij} C_{ij}^{(c)}$, yielding a per-class dissimilarity cost $s_{AB}^{(c)}$. These per-class costs are then aggregated (e.g., via weighted averaging based on class prevalence) and normalized to produce the final cost metric $\tilde{s}_{AB}$. The component $d_S(\cdot,\cdot)$ is a metric on the activation sphere, and $H(\cdot,\cdot)$ is a metric for distributions; for $w_f,w_l>0$, $C_{ij}^{(c)}$ is a valid cost for optimal transport. The aggregated and normalized cost $\tilde{s}_{AB}$ then reflects overall per-class heterogeneity.

\section{Experiments}

We evaluate our metric across diverse FL scenarios, comparing its predictive power against standard Wasserstein distance. Our experiments span synthetic benchmarks with controlled heterogeneity and real-world medical imaging tasks with natural distribution shifts.

\subsection{Experimental Setup}

\paragraph{Datasets and Heterogeneity} 
We systematically evaluate six datasets (Table \ref{table:dataset_summary}). For synthetic and benchmark datasets (Credit, EMNIST, CIFAR), we introduce controlled heterogeneity through three mechanisms: feature skew modifies marginal distributions $p(X)$ by applying site-specific transformations, label skew creates imbalanced class distributions $p(y)$ using Dirichlet allocation, and concept shift alters conditional distributions $p(y|X)$ through label permutation. The medical imaging datasets (IXITiny, ISIC2019) were collected from different sites, thus exhibit natural heterogeneity arising from differences in acquisition protocols, patient populations, and annotation practices across sites. Implementation details for heterogeneity generation are available in our repository~\cite{ourrepo}.

\paragraph{Training}
We compare five training strategies to assess when federation benefits performance over single-site/local training: (1) \textbf{Local training} serves as our baseline, with each site training independently and results averaged across sites. (2) \textbf{FedAvg} \cite{mcmahan2017communication} represents standard federated averaging. (3-5) \textbf{FedProx} \cite{li2020federated}, \textbf{pFedME} \cite{t2020personalized}, and \textbf{Ditto} \cite{li2021ditto} are personalized FL methods designed to handle non-IID data through regularization, dual optimization, and multi-task learning respectively. This selection allows us to evaluate the effect of personalized FL approaches under varying heterogeneity levels.

\begin{table}[h]
\centering
\small
\begin{tabularx}{\columnwidth}{lXc}
\toprule
\textbf{Dataset} & \textbf{Task Type} & \textbf{Heterogeneity} \\
\midrule
Synthetic & Binary classification & Controlled: F, L, C \\
Credit & Fraud detection & Controlled: F, L, C \\
EMNIST & Character recognition & Controlled: F, L, C \\
CIFAR & Object classification & Controlled: F, L, C \\
IXITiny* & Brain MRI segmentation & Natural: site-specific \\
ISIC2019* & Skin lesion diagnosis & Natural: site-specific \\
\bottomrule
\end{tabularx}
\caption{Dataset characteristics. F: feature skew, L: label skew, C: concept shift. *Multi-site medical imaging datasets.}
\label{table:dataset_summary}
\end{table}
\subsection{Synthetic Dataset}

\begin{figure*}
\centering
\makebox[\linewidth][c]{  
  \begin{minipage}{.3\linewidth}
    \centering
    \raisebox{-\height}{\includegraphics[width=\linewidth]{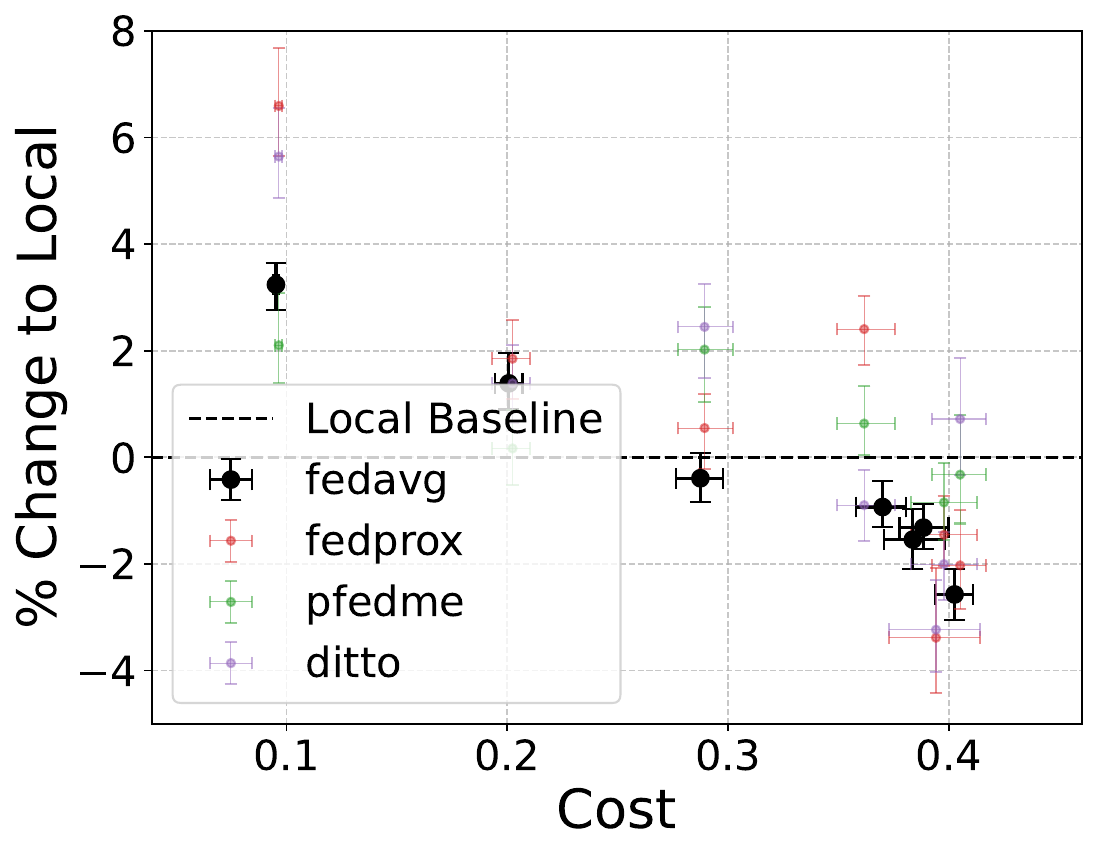}}
    \makebox[\linewidth][c]{\small\textit{Synthetic Feature}} 
  \end{minipage}%
    \begin{minipage}{.3\linewidth}
    \centering
    \raisebox{-\height}{\includegraphics[width=\linewidth]{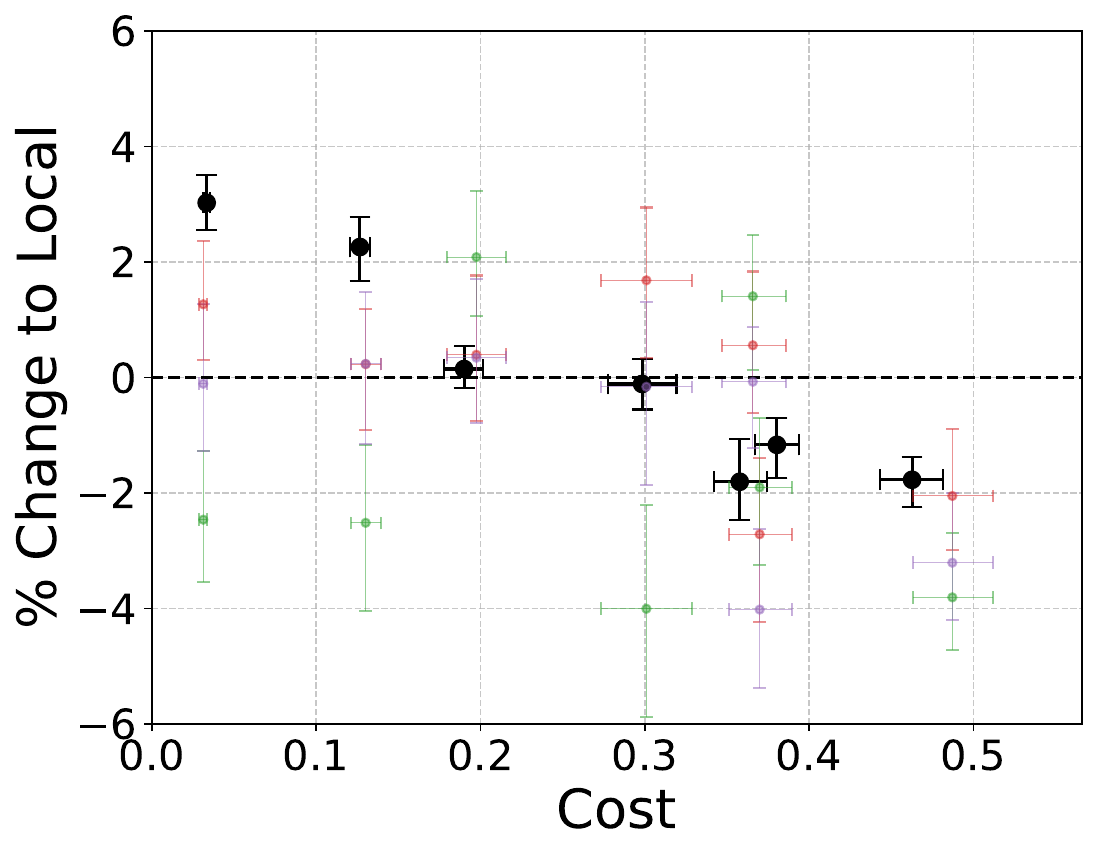}}
    \makebox[\linewidth][c]{\small\textit{Credit}} 
  \end{minipage}%
  \begin{minipage}{.3\linewidth}
    \centering
    \raisebox{-\height}{\includegraphics[width=\linewidth]{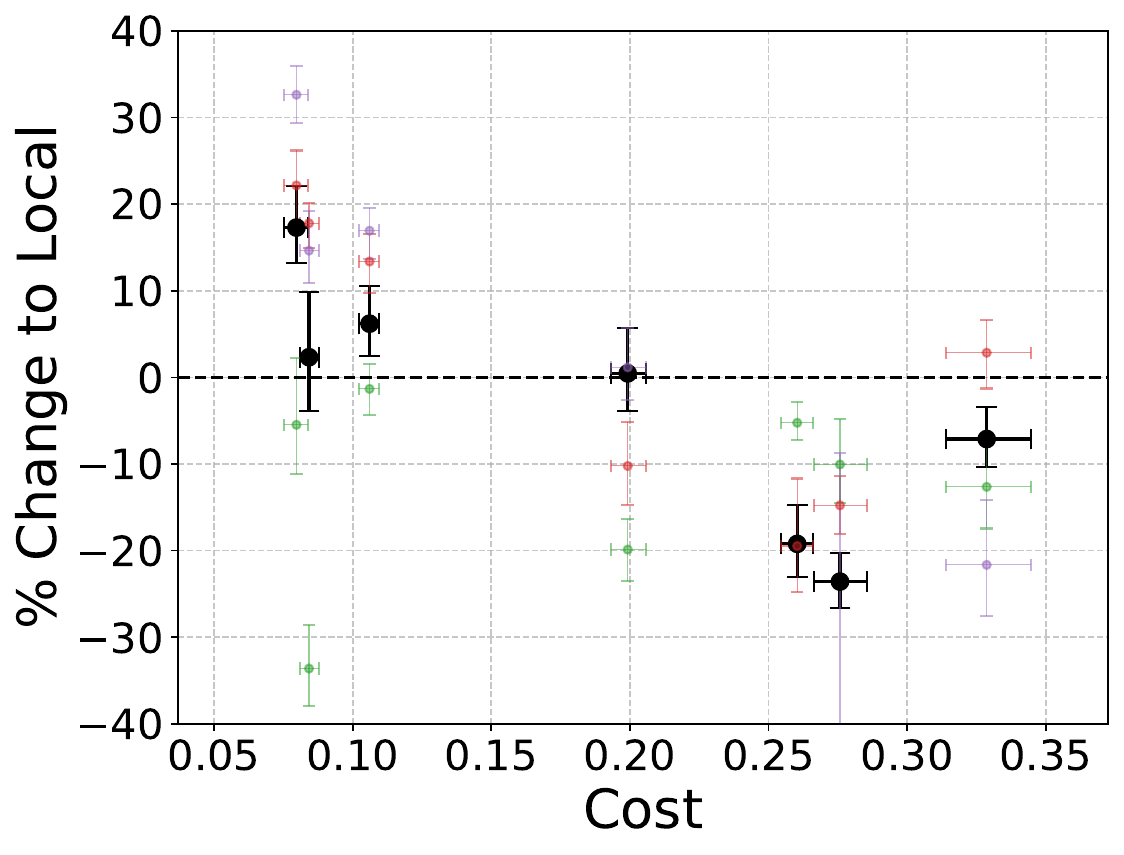}}
    \makebox[\linewidth][c]{\small\textit{EMNIST}} 
  \end{minipage}
}
\par\vspace{0.5cm}
\makebox[\linewidth][c]{  
  \begin{minipage}{.3\linewidth}
    \centering
    \raisebox{-\height}{\includegraphics[width=\linewidth]{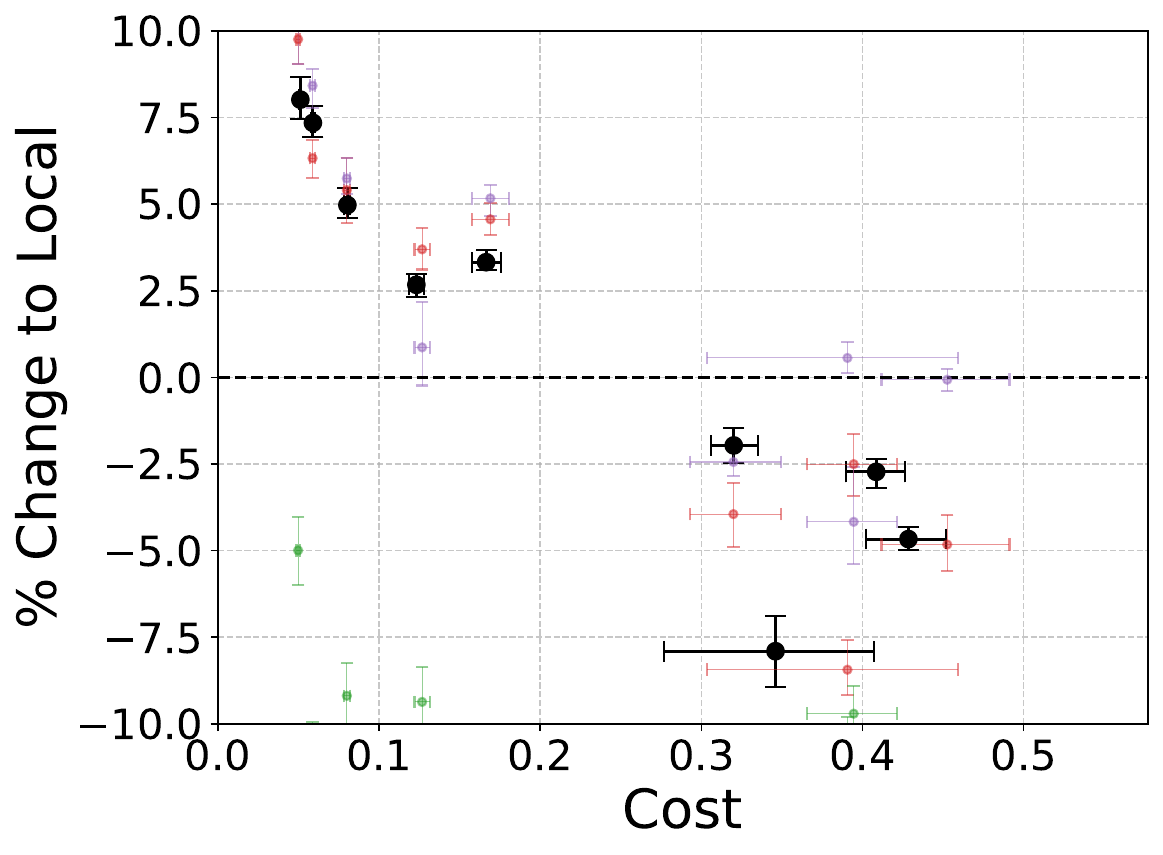}}
    \makebox[\linewidth][c]{\small\textit{CIFAR-100}} 
  \end{minipage}%
  \begin{minipage}{.3\linewidth}
    \centering
    \raisebox{-\height}{\includegraphics[width=\linewidth]{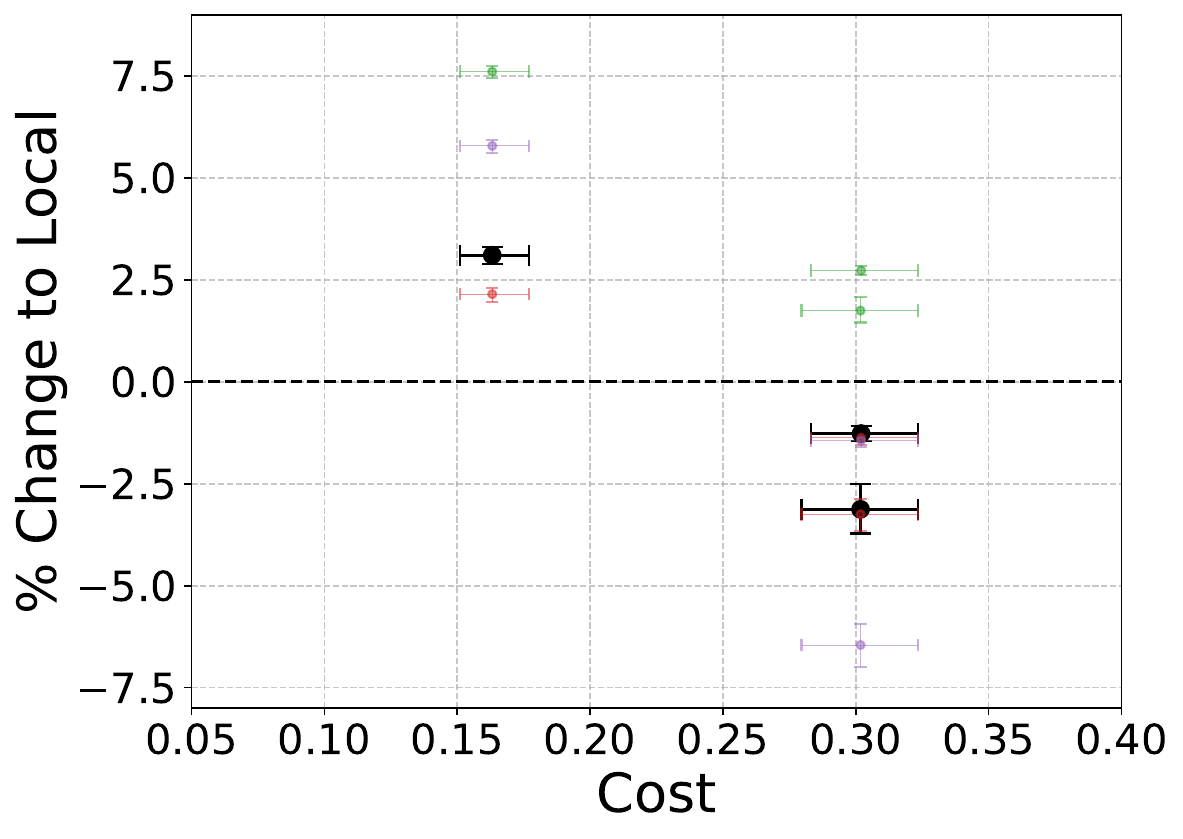}}
    \makebox[\linewidth][c]{\small\textit{IXITiny}} 
  \end{minipage}%
  \begin{minipage}{.3\linewidth}
    \centering
    \raisebox{-\height}{\includegraphics[width=\linewidth]{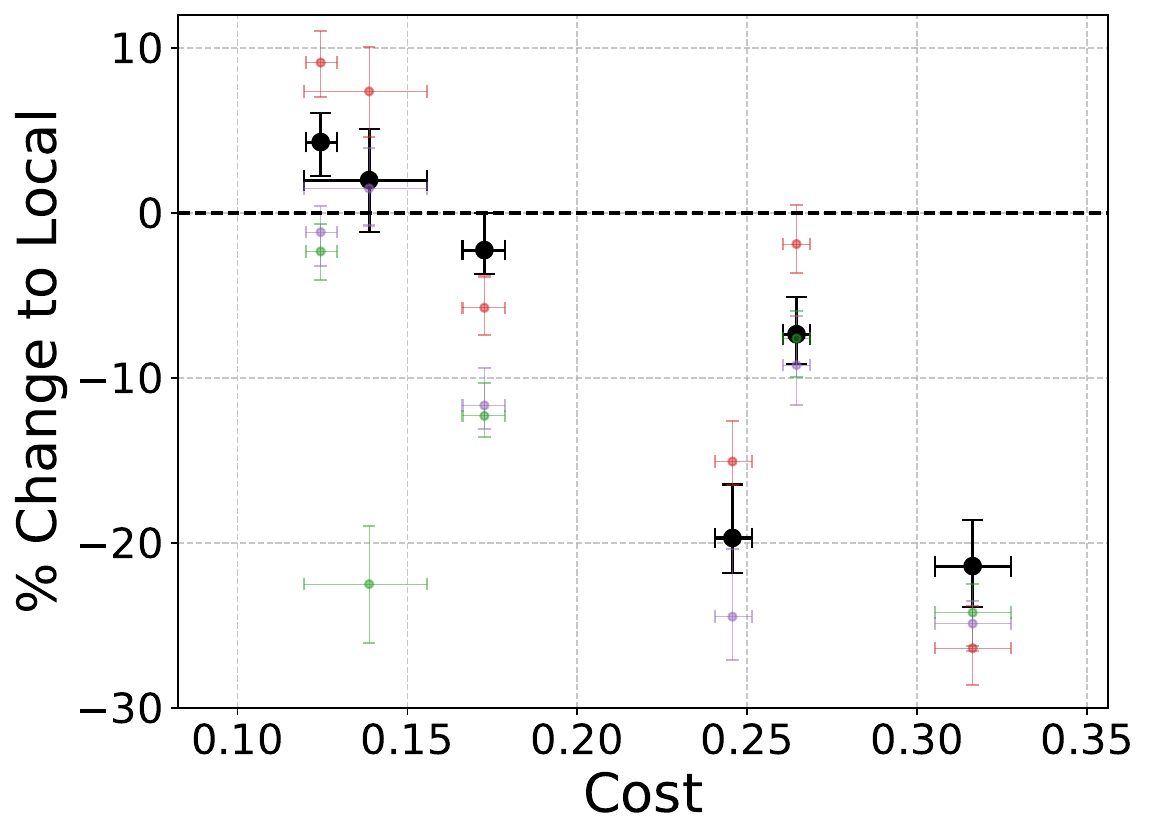}}
    \makebox[\linewidth][c]{\small\textit{ISIC-2019}} 
  \end{minipage}
}
\caption{\textbf{Performance across varying costs}. Results show percentage improvement over local training baseline, with FedAvg (black) as the primary comparison. FedProx (red), pFedME (green), and Ditto (purple) are shown with reduced opacity for reference.}
\label{fig:performance}
\end{figure*}

To gain an understanding of our metric's behavior, we evaluate it on a synthetic dataset. This allows us to control distribution shifts and evaluate a range of scenarios. Our approach involves drawing samples from two distinct distributions, varying how much overlap in distributions the datasets share. We find that when  datasets are drawn from the same distribution, we obtain a cost of $\leq0.1$ and find an improvement in model performance using FL (Figure \ref{fig:performance}) and when datasets are drawn from distinct distributions, we obtain a cost of $\geq0.3$ and find worsened performance using FL. We find that as long as costs are $\leq0.2$, FL improves performance. As discussed in Section \ref{subsec:activation_dissimilarity}, this is not surprising as we expect vectors to be orthogonal when they are unrelated.

\subsection{Benchmark Datasets}

Next, we evaluate the metric on benchmark datasets: Credit, EMNIST, and CIFAR-10. This allows us to replicate the models and distribution shifts presented in many FL studies in the literature. Our findings confirm observations from the synthetic dataset; \textbf{costs $\leq 0.2$ generally result in improved FL performance, while costs $\geq 0.3$ lead to a decline in model performance} compared to the baseline (Figure \ref{fig:performance}). This finding is consistent across all datasets and distribution shift types, highlighting the robustness of our metric. 

\subsection{Real-world Datasets}
To validate our metric's ability to capture \textbf{real-world differences} across datasets, we tested it on two medical imaging datasets: IXITiny, an image segmentation task on 3D brain MRIs, and ISIC2019, a skin lesion classification task using dermoscopy images. Both datasets have \textbf{natural non-IID partitions} as they are made up of samples collected from different sites using different machines. We compared performance across pairs of sites with differing metric costs. We obtain results consistent with other synthetic datasets.

As both datasets have real-world partitions, we compare our results to what is known.
\begin{itemize}

    \item \textbf{IXITiny}: The costs produced by our metric aligned with the data collection practices. Two sites used similar Philips MRI machines with comparable imaging parameters, while the third site employed a different GE machine with unspecified settings \cite{ixi_dataset}. The metric assigned a low cost of 0.08 between the two Philips sites and higher costs of 0.28 and 0.30 between the GE site and the other two, respectively. This is consistent with \cite{terrail2022flamby}.

    \item \textbf{ISIC2019}: Our cost aligns with how the data was collected while providing some novel insights. ISIC2019 data is collected from four sites (two in Europe, one in USA, one in Australia). One of the European sites used three different machines, resulting in separate data partitions. We find that European sites have lower costs with each other than with Australian and American sites, consistent with \cite{terrail2022flamby}. Interestingly, we find that, among European sites, imaging machine has more impact on dataset similarity and \textit{model performance} than geographical location. \cite{wen2022characteristics} discuss a similar observation.
\end{itemize} 

Although the differences in these datasets are well-documented, our metric offers significant utility when they are not known. This information can guide the selection of sub-networks of sites with similar datasets for collaboration, potentially leading to improved FL performance. Additionally, in cases like medical imaging, the metric can inform the extent of data pre-processing steps needed to normalize samples and mitigate the negative effects of non-IID data.

Through our experiments with both datasets, we observed moderate variations in sample sizes across sites, with some locations having two to three times more samples than others. Despite these differences, our metric maintained consistent and reliable interpretations, demonstrating its robustness to moderate sample size variations.

\subsection{Impact of Personalized FL Algorithms}
\begin{figure*}[!ht]

\centering

\makebox[\linewidth][c]{
  \begin{minipage}{.9\linewidth}
    \centering
    \raisebox{-\height}{\includegraphics[width=\linewidth]{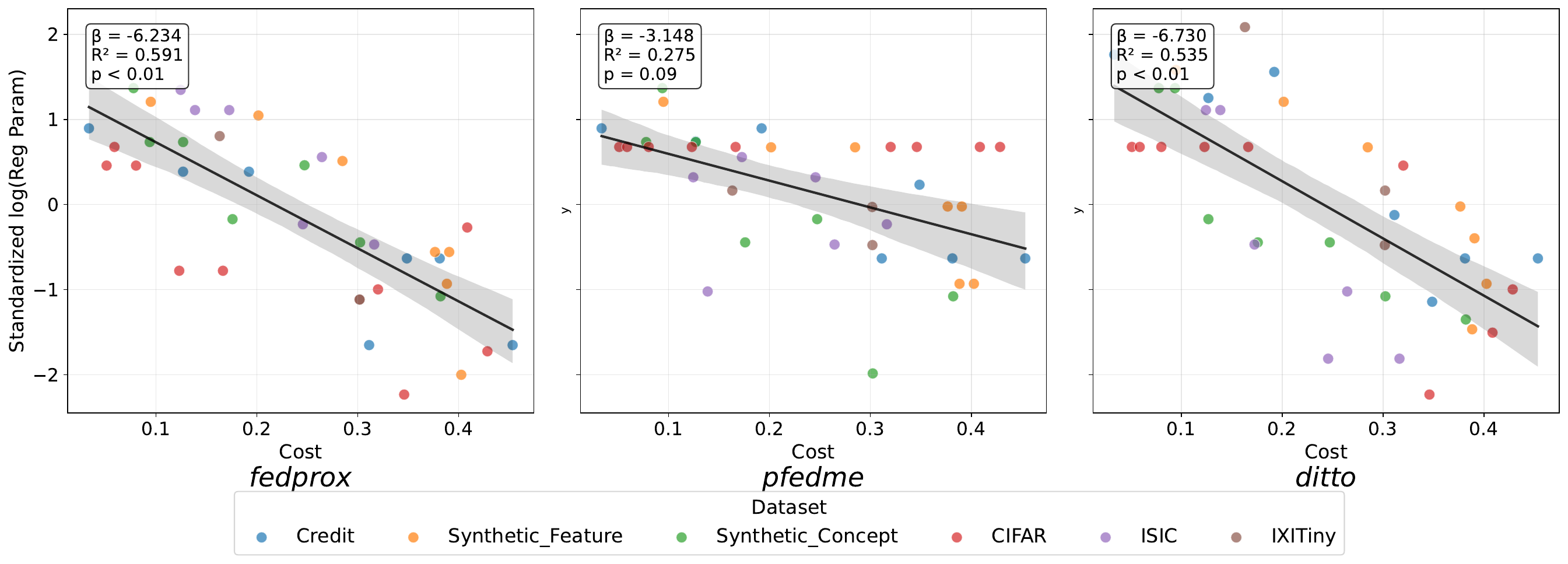}}
  \end{minipage}%
}
\caption{\textbf{Relationship between optimal regularization parameters and metric for personalized FL algorithms.} Higher costs correlate with lower parameters, indicating stronger personalization is beneficial when clients have heterogeneous data.}
\label{fig:pfl_reg_param}

\end{figure*}

Personalized FL algorithms—FedProx, pFedME, and Ditto—can mitigate performance degradation from high inter-client dataset dissimilarity, but their effectiveness depends critically on proper regularization tuning. These algorithms incorporate regularization terms that control how much local client models can deviate from the global model, with pFedME and Ditto additionally maintaining separate personalized models for each client. To investigate whether our metric can guide regularization selection, we conducted comprehensive hyperparameter searches across regularization values from $10^{-6}$ to $5$ for each algorithm-dataset combination. We then analyzed the relationship between optimal regularization parameters and our metric. To enable robust comparison across different experimental settings and numerical scales, regularization values were log-transformed and standardized. Figure~\ref{fig:pfl_reg_param} reveals a statistically significant negative relationship between our metric and optimal regularization strength for FedProx ($\beta = -6.234$, $p < 0.01$) and Ditto ($\beta = -6.730$, $p < 0.01$) with a similar trend, though not statistically significant, for pFedME  ($\beta = -3.148$, $p = 0.09$). This indicates that as dataset dissimilarity increases (higher cost), weaker regularization becomes optimal, allowing stronger personalization by reducing the global model's influence on local updates.

These findings demonstrate that our metric provides actionable guidance for personalized FL deployment: higher dissimilarity costs suggest favoring personalization over federation, while lower costs indicate that standard FedAvg may suffice. This relationship enables practitioners to make principled algorithmic choices based on dataset characteristics rather than extensive empirical tuning.

\subsection{Relationship to Weight Divergence}
\begin{figure*}[!ht]
\centering

\makebox[\linewidth][c]{  
\begin{minipage}{.3\linewidth}
\centering
\raisebox{-\height}{\includegraphics[width=\linewidth]{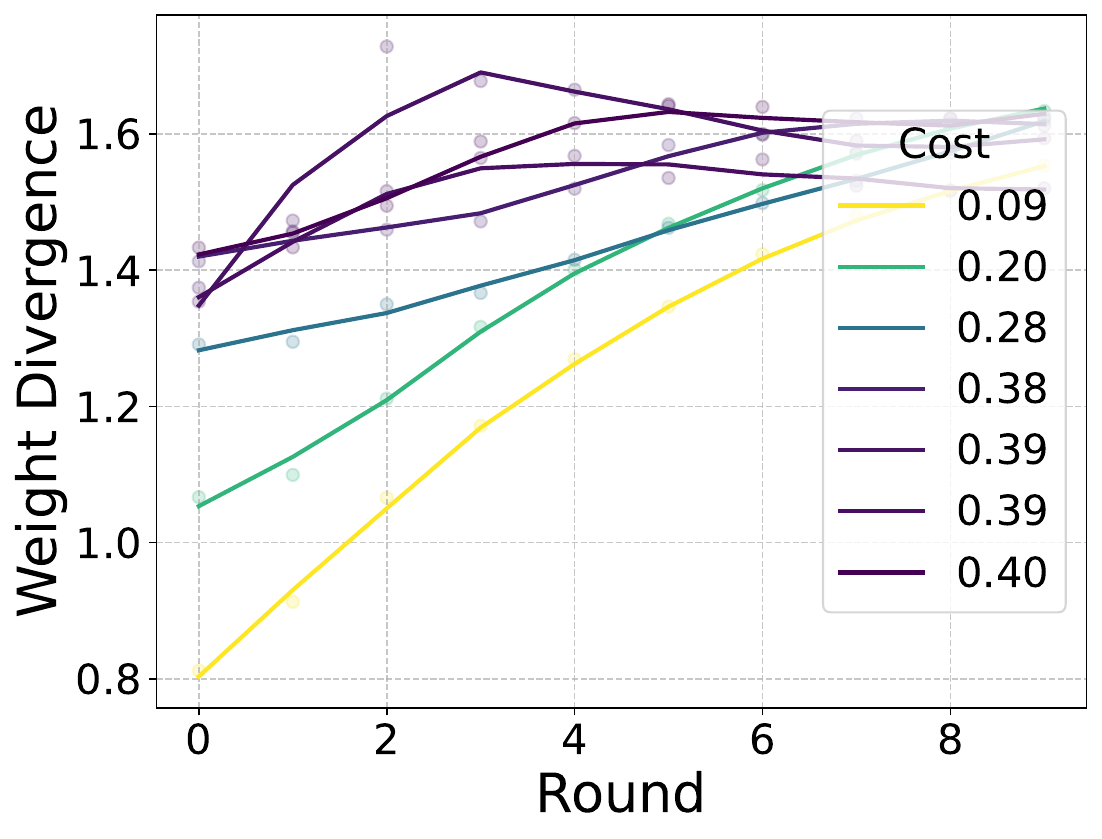}}
\makebox[\linewidth][c]{\small\textit{Synthetic Feature}} 
\end{minipage}%
  \begin{minipage}{.3\linewidth}
    \centering
    \raisebox{-\height}{\includegraphics[width=\linewidth]{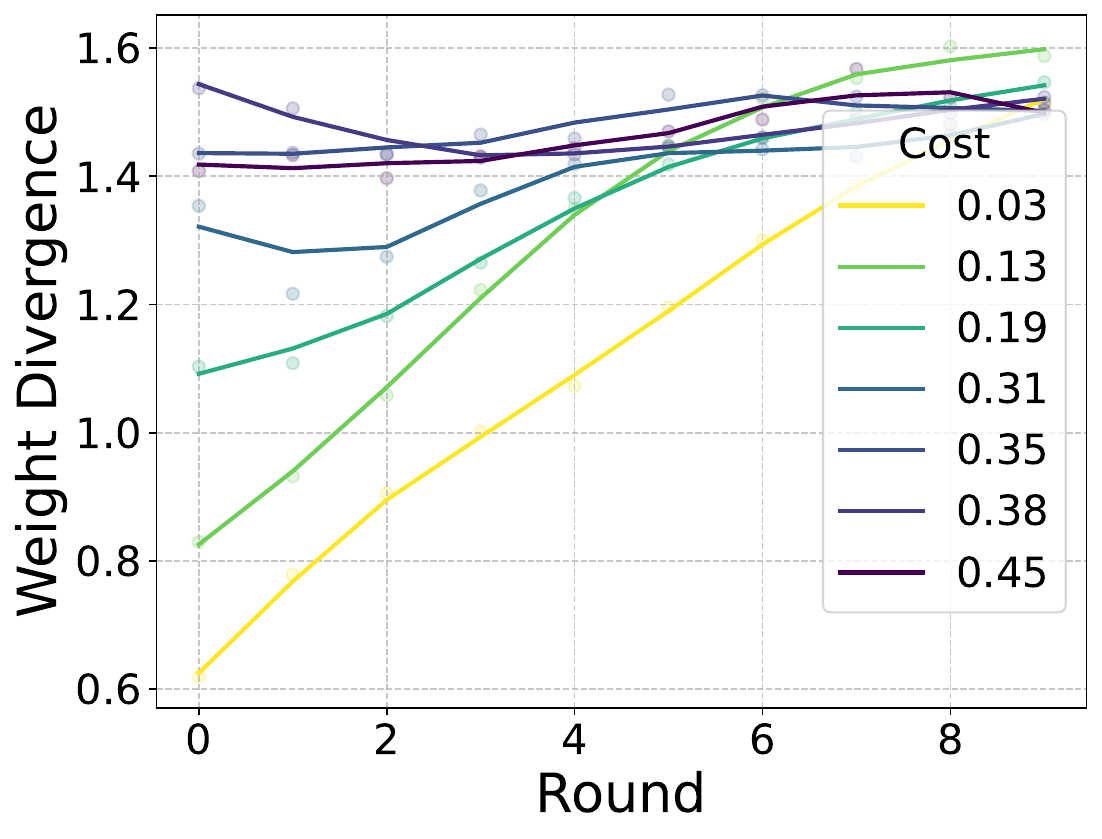}}
    \makebox[\linewidth][c]{\small\textit{Credit}} 
  \end{minipage}%
  \begin{minipage}{.3\linewidth}
    \centering
    \raisebox{-\height}{\includegraphics[width=\linewidth]{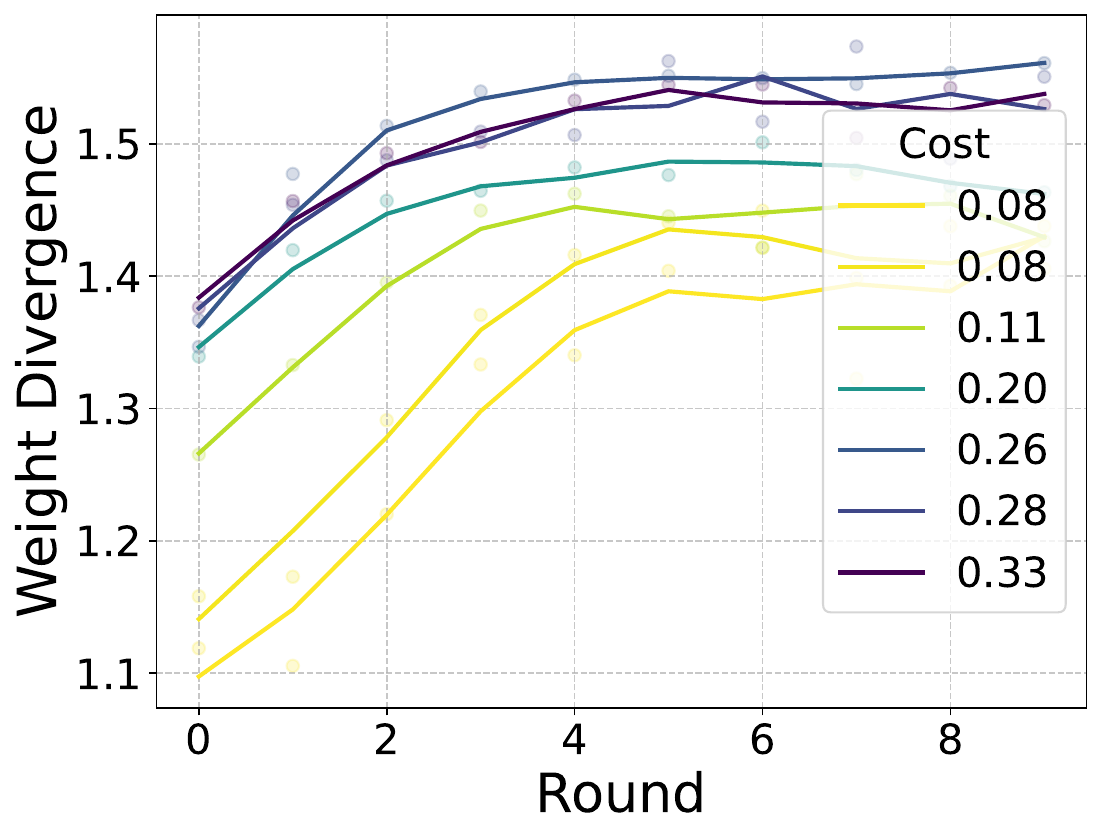}}
    \makebox[\linewidth][c]{\small\textit{EMNIST}} 
  \end{minipage}
}
\par\vspace{0.5cm}
\makebox[\linewidth][c]{  
  \begin{minipage}{.3\linewidth}
    \centering
    \raisebox{-\height}{\includegraphics[width=\linewidth]{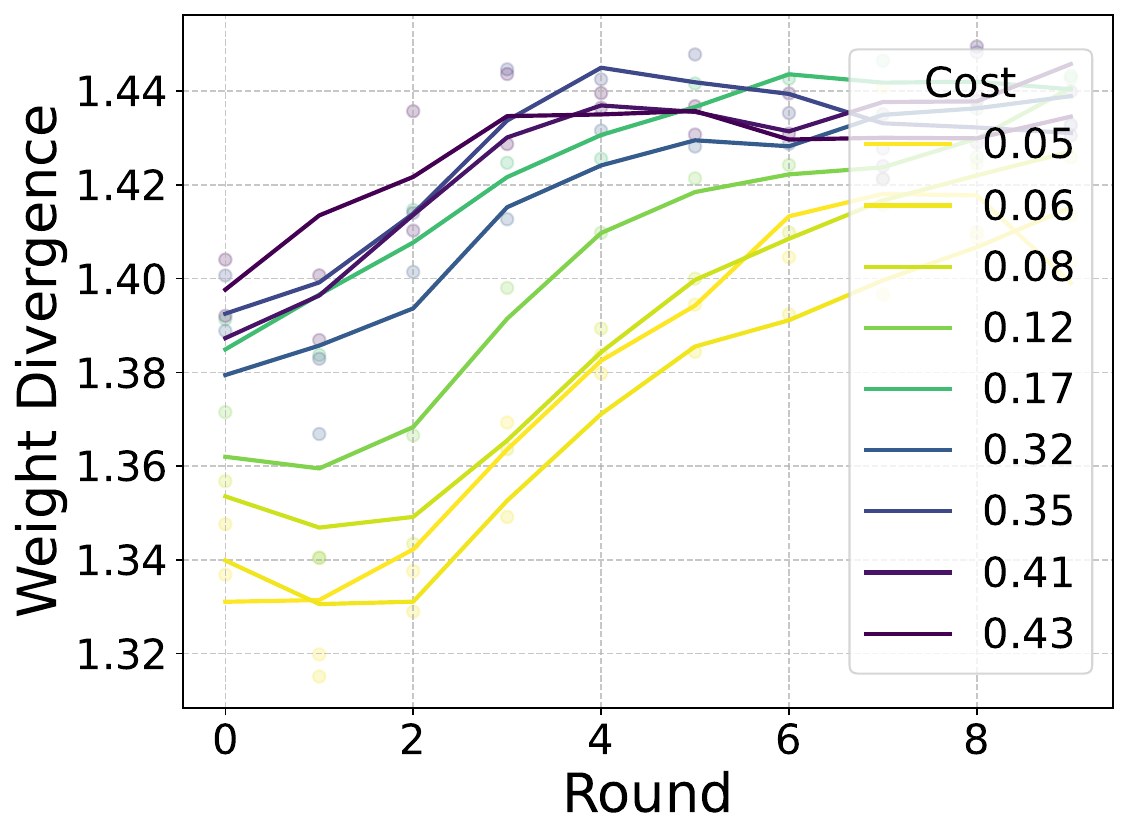}}
    \makebox[\linewidth][c]{\small\textit{CIFAR}} 
  \end{minipage}%
  \begin{minipage}{.3\linewidth}
    \centering
    \raisebox{-\height}{\includegraphics[width=\linewidth]{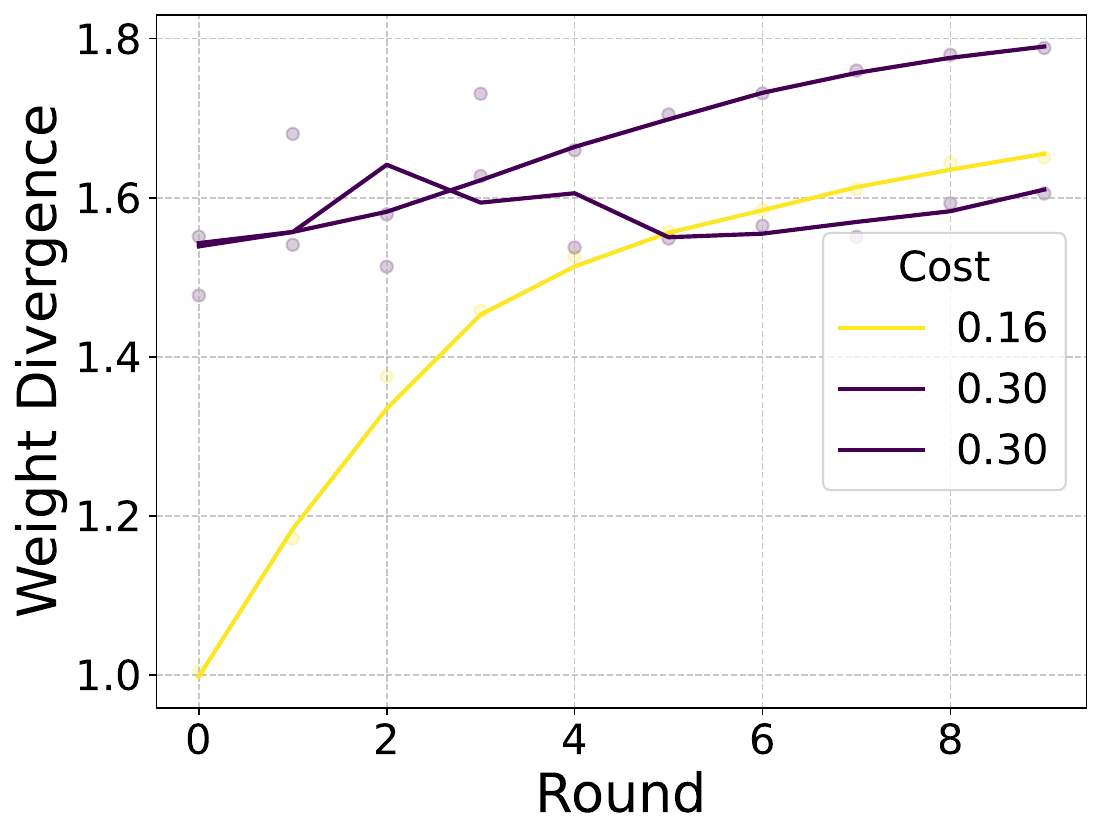}}
    \makebox[\linewidth][c]{\small\textit{IXITiny}} 
  \end{minipage}%
  \begin{minipage}{.3\linewidth}
    \centering
    \raisebox{-\height}{\includegraphics[width=\linewidth]{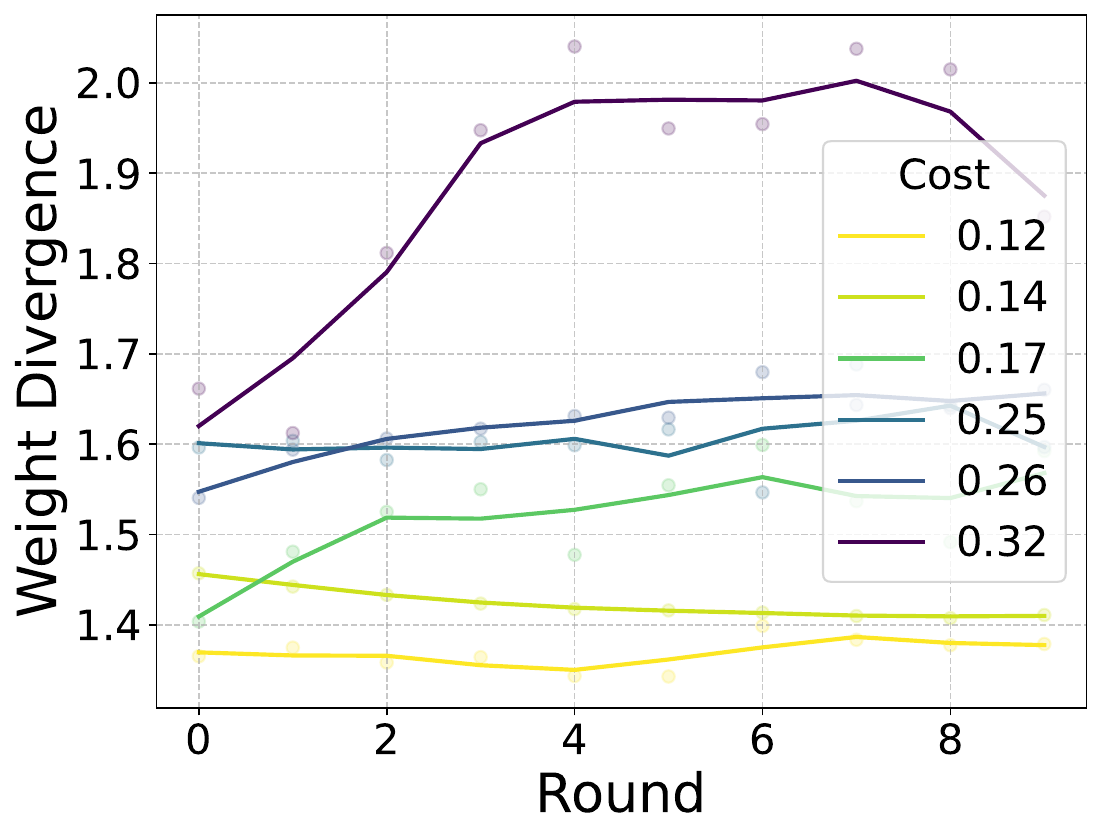}}
    \makebox[\linewidth][c]{\small\textit{ISIC-2019}} 
  \end{minipage}
}

\caption{\textbf{Weight divergence during training for FedAvg models}}
\label{fig:weight_divergence}
\end{figure*}
\label{subsec:weight_div}
To validate the connection between our metric and FL performance, we tracked weight divergence (Equation \ref{eqn:weight_div}) during training with FedAvg (Figure \ref{fig:weight_divergence}). Higher metric costs consistently predict greater weight divergence, with this divergence manifesting earlier in training. Our metric, calculated after a single federated learning round, reliably predicts divergence behavior across the complete training process. This confirms that analyzing model activations from the initial round provides sufficient information for practitioners to make well-informed algorithmic choices without needing to complete entire training cycles. Our metric significantly outperforms standard Wasserstein distance in predicting weight divergence (paired t-test, $p=0.013$), providing validation for our approach. High costs capture fundamental misalignments in how clients represent features and encode classes, which manifest as persistent weight divergence during federated optimization.

\subsection{Sample Size Complexity}
\label{sec:sample_size}
\begin{figure*}[!ht]

\centering

\makebox[\linewidth][c]{  
  \begin{minipage}{.27\linewidth}
    \centering
    \raisebox{-\height}{\includegraphics[width=\linewidth]{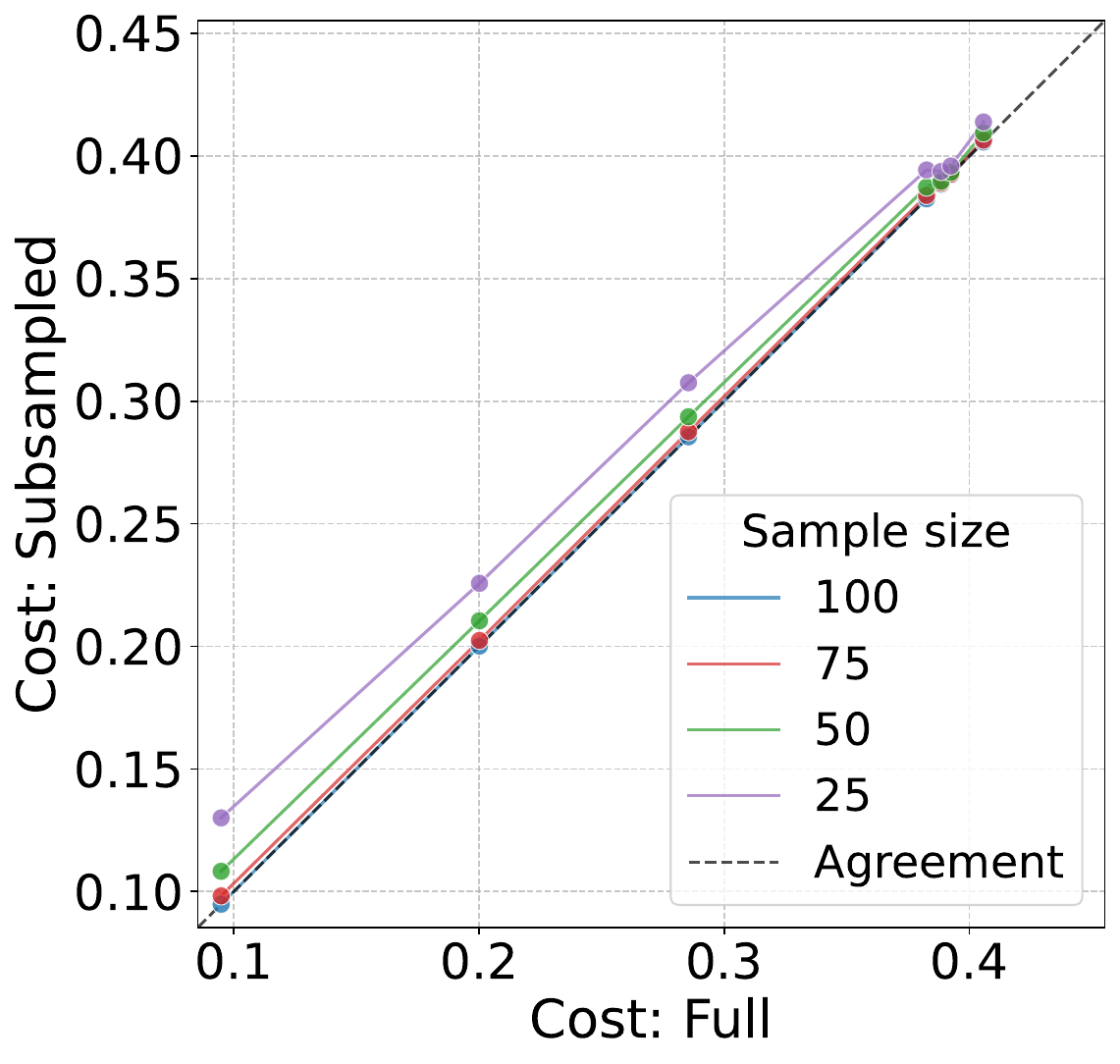}}
    \makebox[\linewidth][c]{\small\textit{Synthetic Feature }} 
  \end{minipage}%
  \begin{minipage}{.27\linewidth}
    \centering
    \raisebox{-\height}{\includegraphics[width=\linewidth]{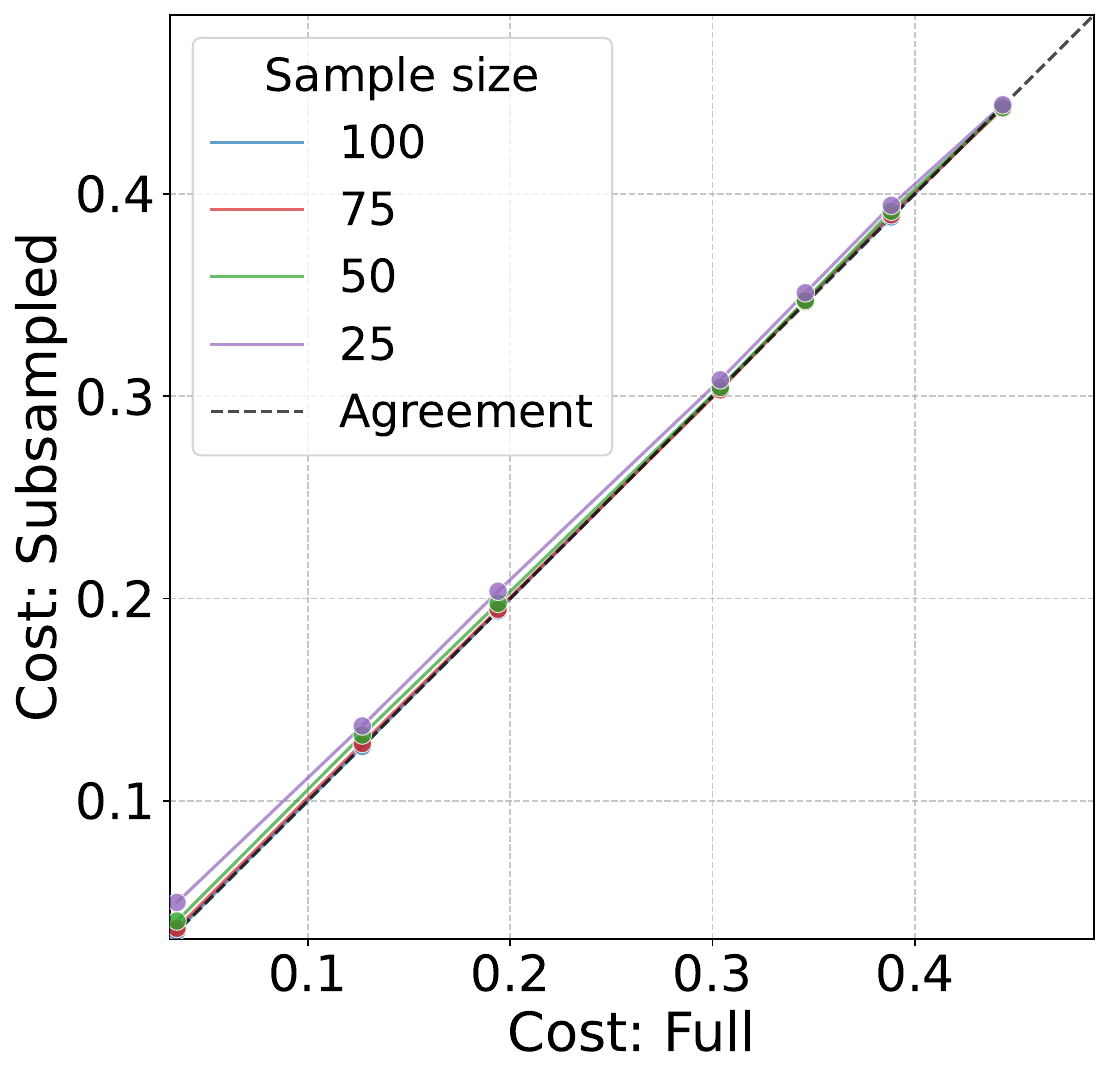}}
    \makebox[\linewidth][c]{\small\textit{Credit}} 
  \end{minipage}%

  \begin{minipage}{.27\linewidth}
    \centering
    \raisebox{-\height}{\includegraphics[width=\linewidth]{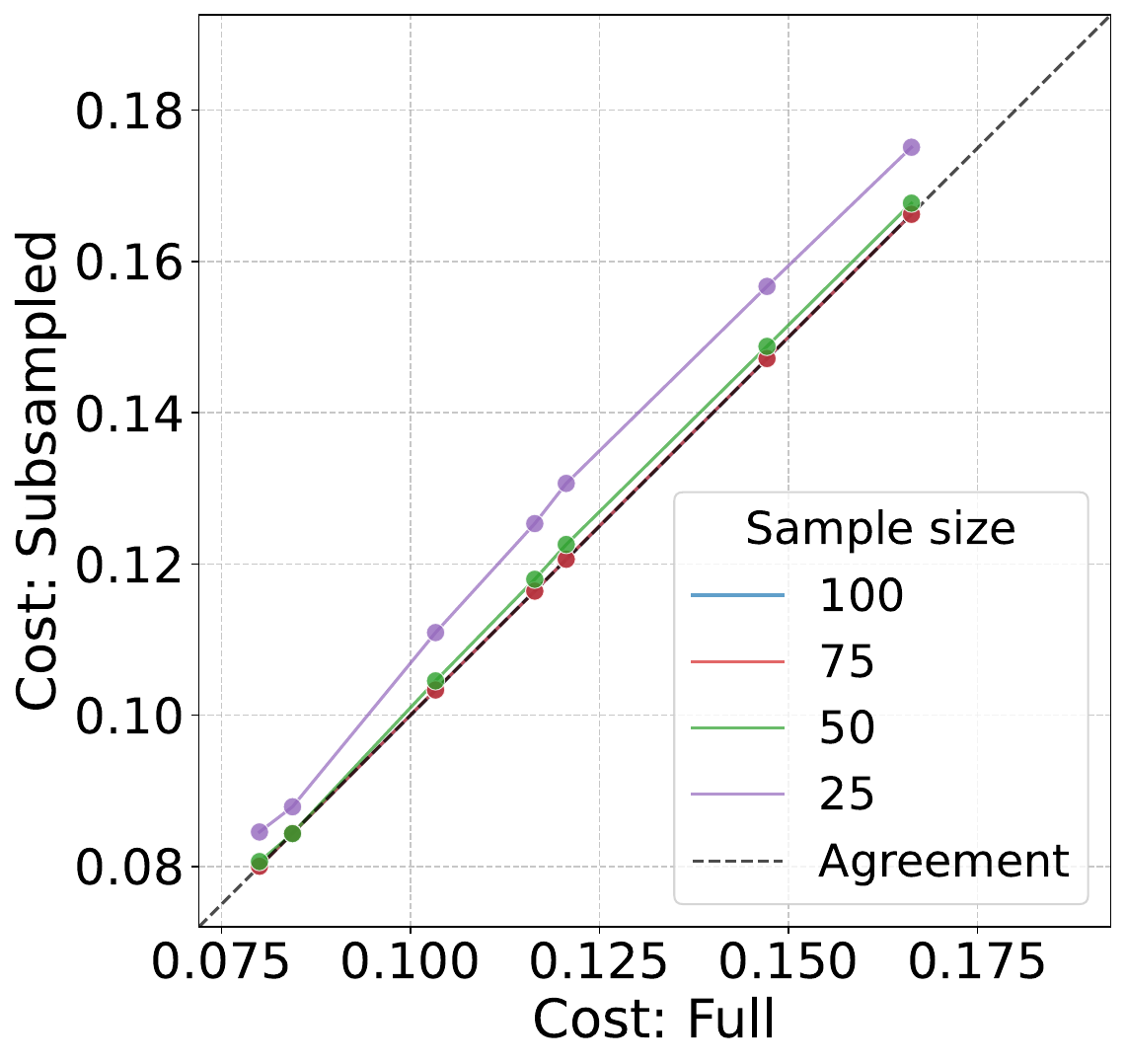}}
    \makebox[\linewidth][c]{\small\textit{EMNIST}} 
  \end{minipage}
}
\par\vspace{0.5cm}
\makebox[\linewidth][c]{  
  \begin{minipage}{.27\linewidth}
    \centering
    \raisebox{-\height}{\includegraphics[width=\linewidth]{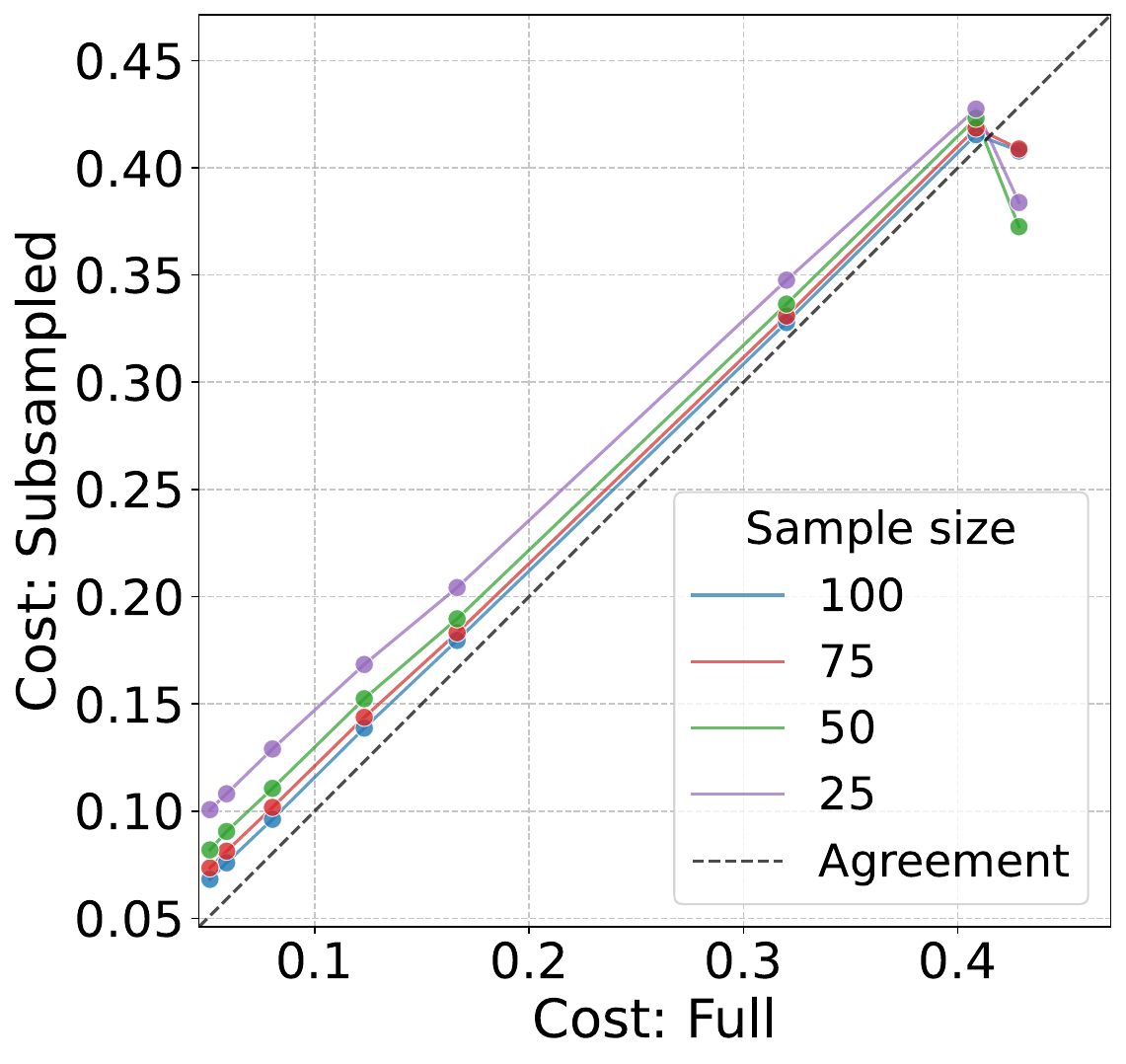}}
    \makebox[\linewidth][c]{\small\textit{CIFAR}} 
  \end{minipage}%
  \begin{minipage}{.27\linewidth}
    \centering
    \raisebox{-\height}{\includegraphics[width=\linewidth]{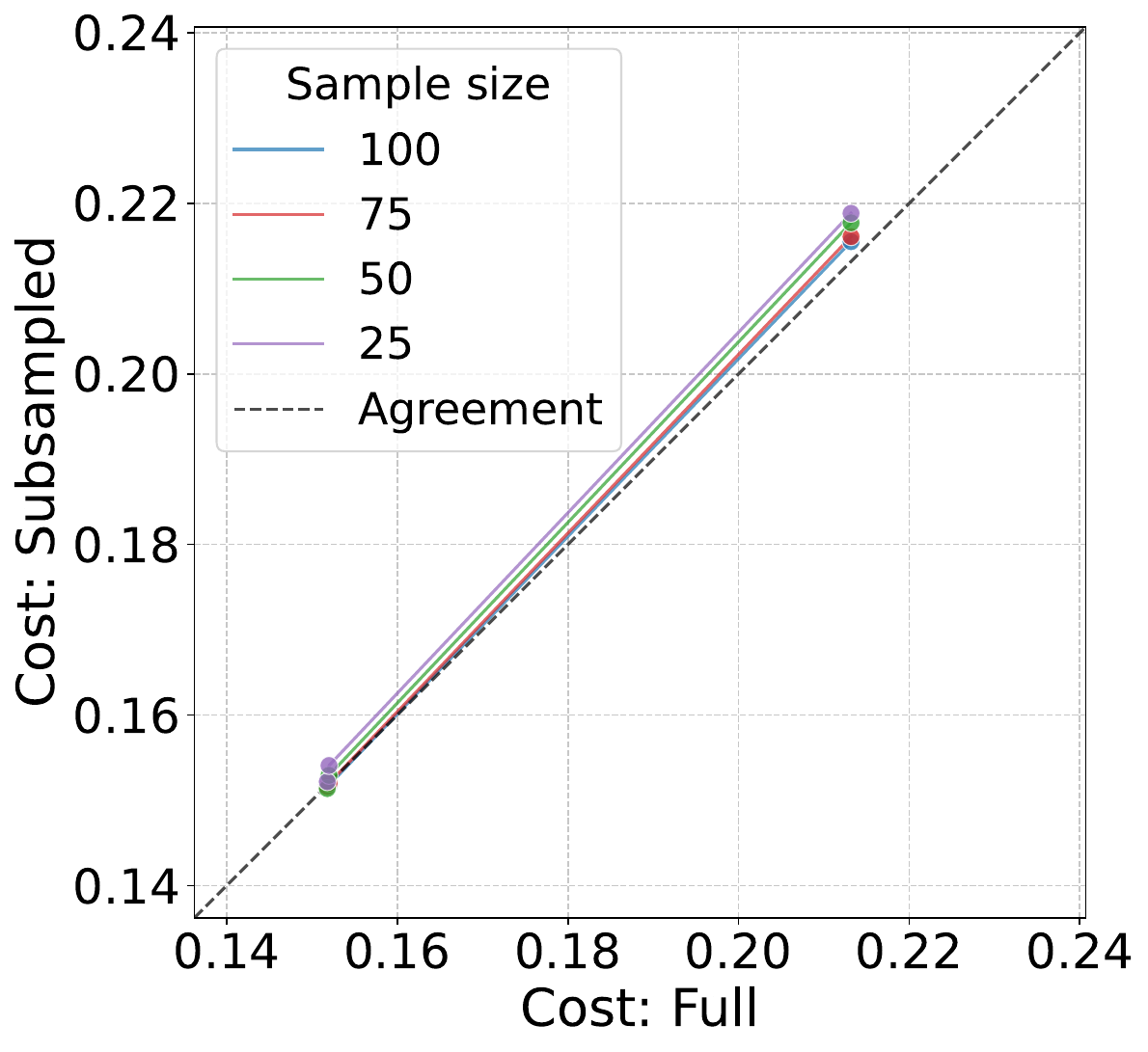}}
    \makebox[\linewidth][c]{\small\textit{IXITiny}} 
  \end{minipage}%
  \begin{minipage}{.27\linewidth}
    \centering
    \raisebox{-\height}{\includegraphics[width=\linewidth]{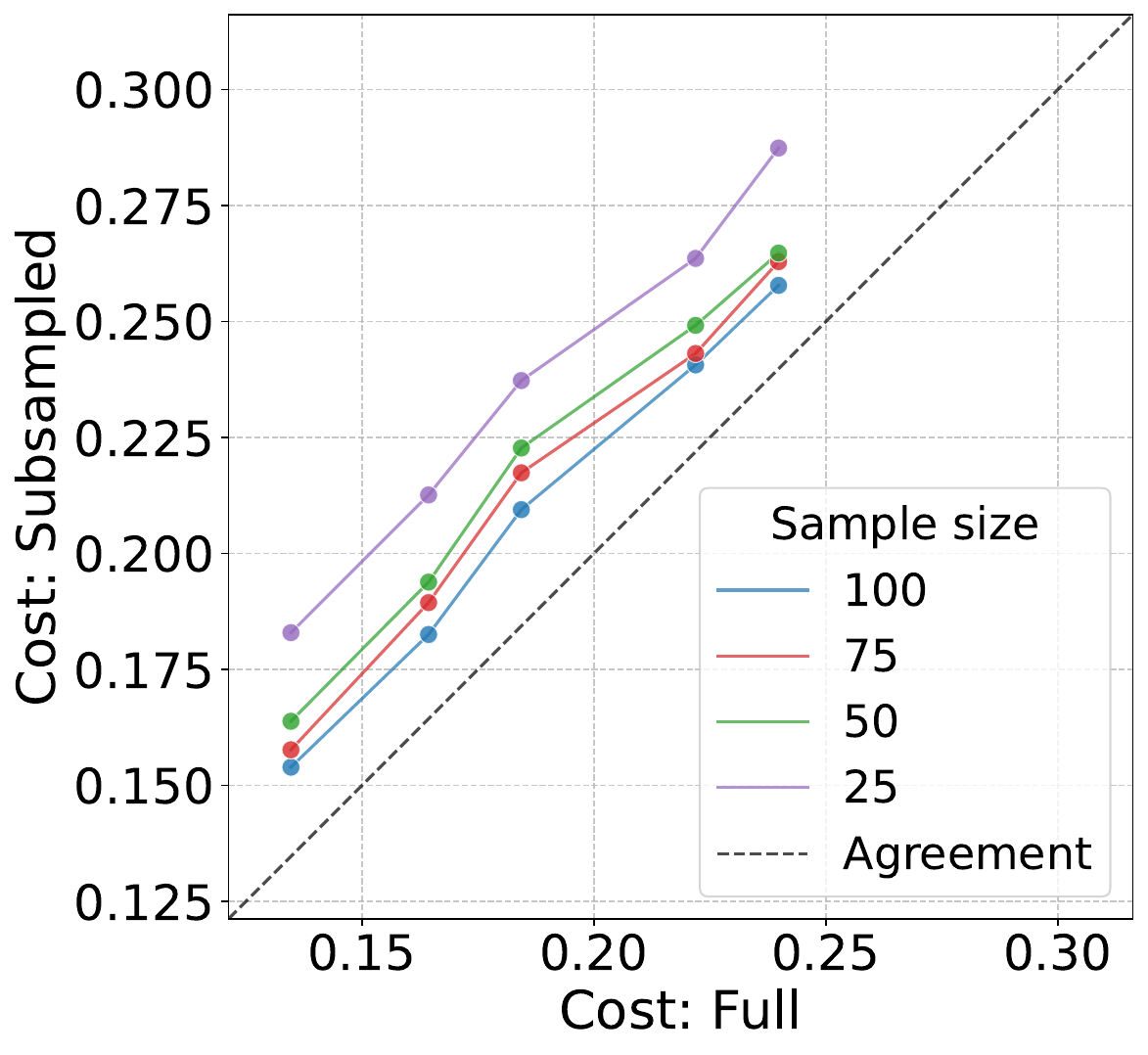}}
    \makebox[\linewidth][c]{\small\textit{ISIC2019}} 
  \end{minipage}
}

\caption{\textbf{Scores calculated from full dataset and subsampled dataset}. Dashed Y=X represents perfect agreement.}
\label{fig:sample_complexity}

\end{figure*}
Our metric is also sample size efficient. This is important in FL, where dataset size in individual sites is limited.  Mena et al., and Genevay et al., \cite{Mena2019, Genevay2019-yt} have already discussed theoretical bounds on sample size efficiency. In Figure \ref{fig:sample_complexity} we present an empirical comparison of the estimated metric costs on the full datasets vs. subsampled datasets. We show that around 50 or more samples are required per label to accurately estimate the cost. Fewer samples than that leads to overestimation.

\subsection{Comparison to Original Wasserstein Distance}
\begin{figure*}[!ht]
\centering

\makebox[\linewidth][c]{  
    \begin{minipage}{.3\linewidth}
    \centering
    \raisebox{-\height}{\includegraphics[width=\linewidth]{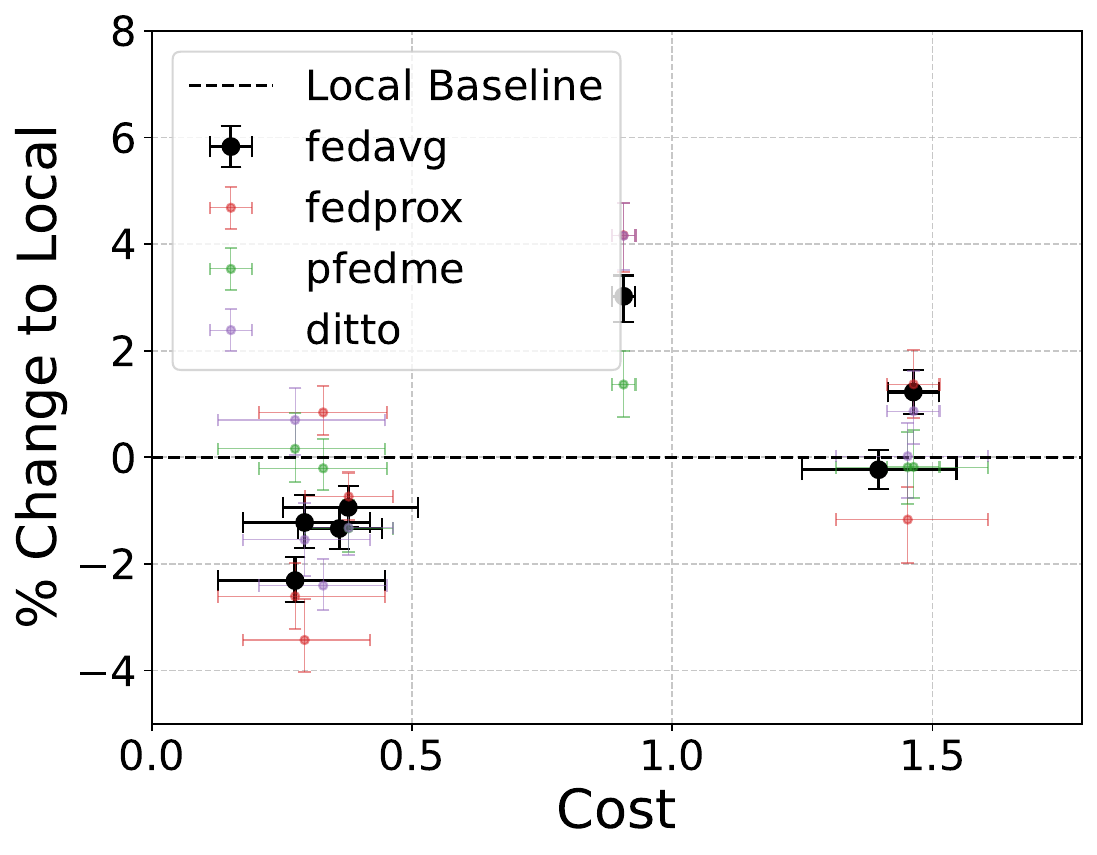}}
    \makebox[\linewidth][c]{\small\textit{Synthetic Feature}} 
  \end{minipage}%
  \begin{minipage}{.3\linewidth}
    \centering
    \raisebox{-\height}{\includegraphics[width=\linewidth]{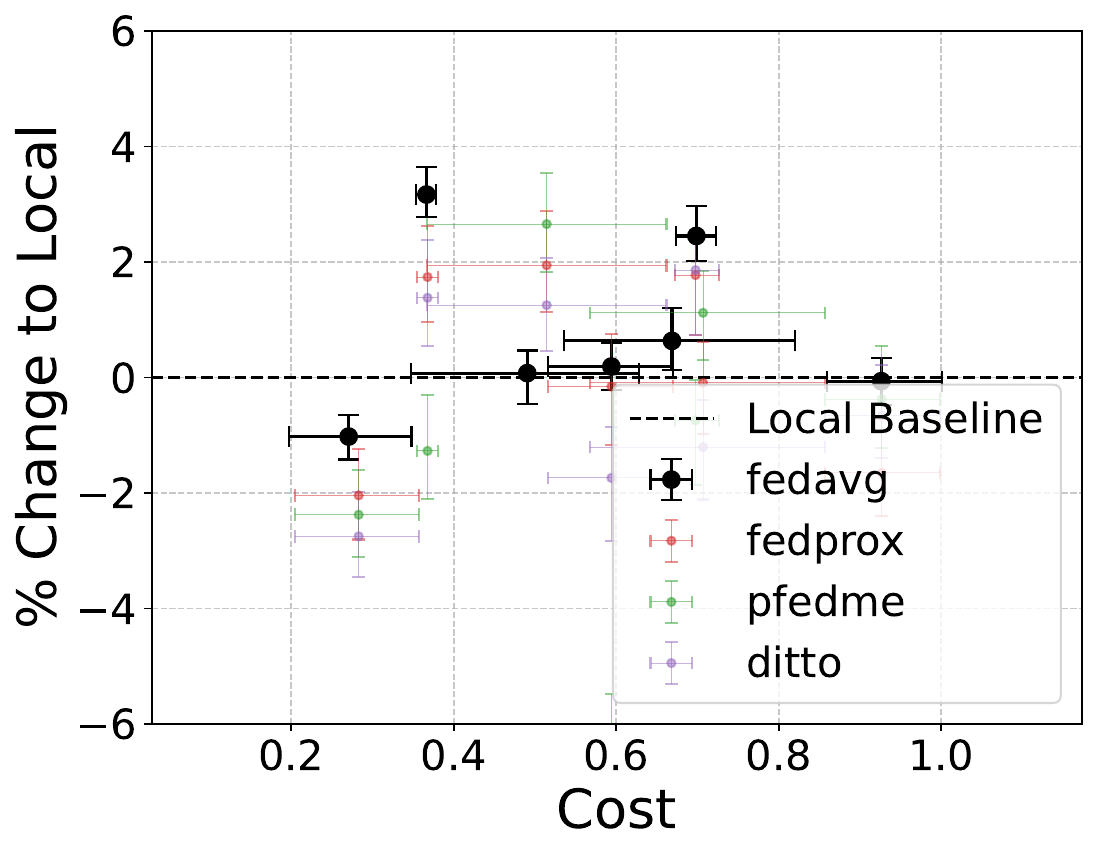}}
    \makebox[\linewidth][c]{\small\textit{Credit}} 
  \end{minipage}%
  \begin{minipage}{.3\linewidth}
    \centering
    \raisebox{-\height}{\includegraphics[width=\linewidth]{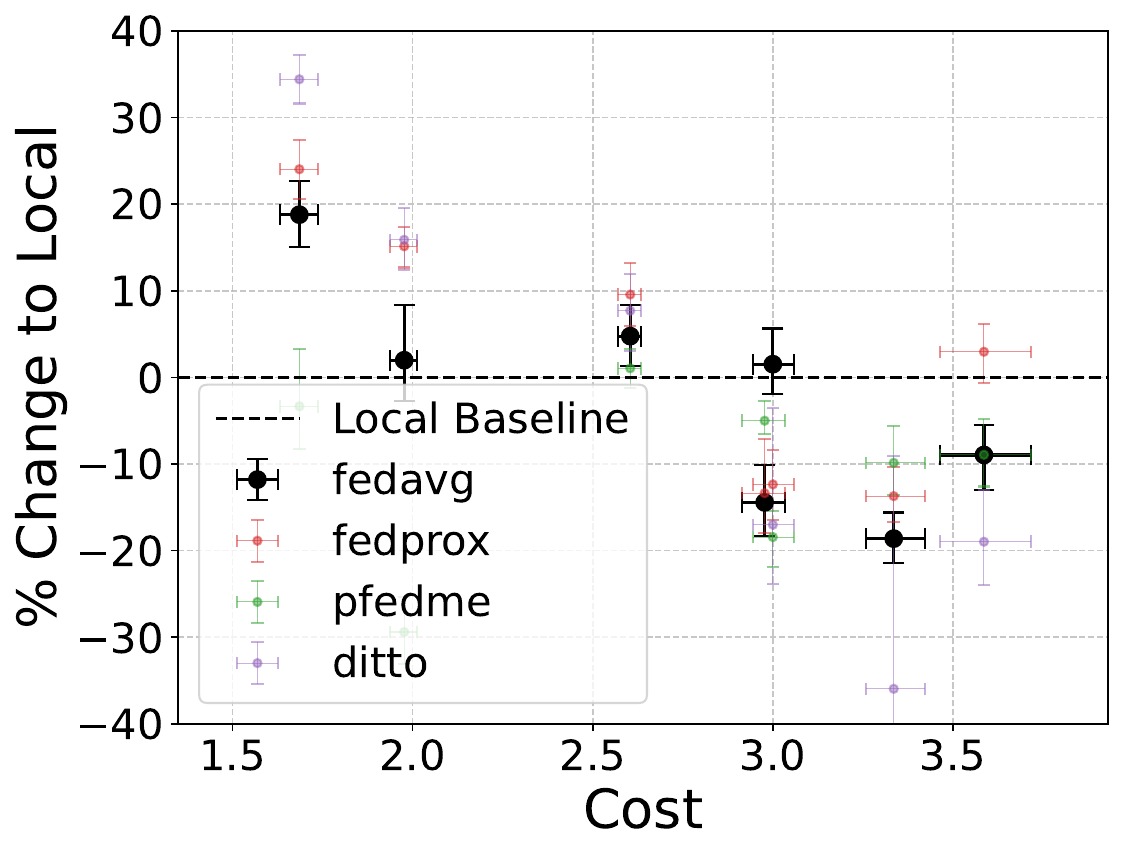}}
    \makebox[\linewidth][c]{\small\textit{EMNIST}} 
  \end{minipage}
}
\par\vspace{0.5cm}
\makebox[\linewidth][c]{  
  \begin{minipage}{.3\linewidth}
    \centering
    \raisebox{-\height}{\includegraphics[width=\linewidth]{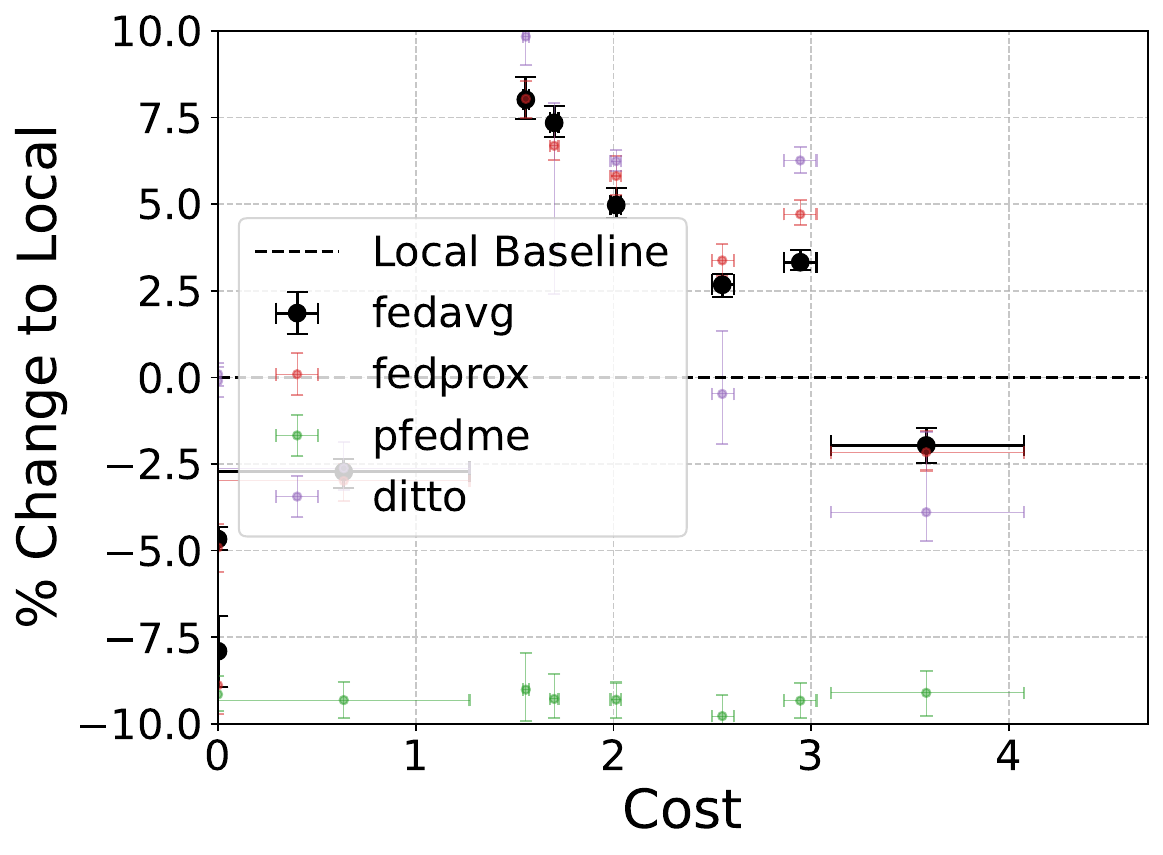}}
    \makebox[\linewidth][c]{\small\textit{CIFAR-100}} 
  \end{minipage}%
  \begin{minipage}{.3\linewidth}
    \centering
    \raisebox{-\height}{\includegraphics[width=\linewidth]{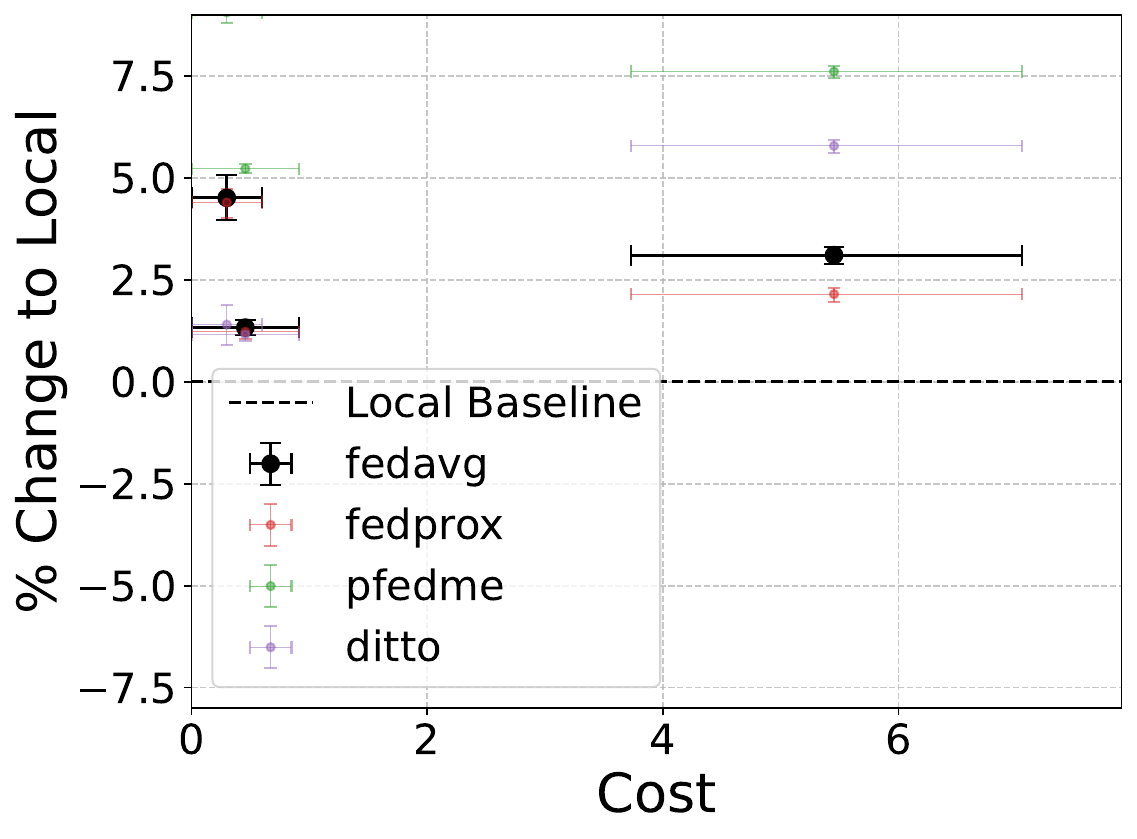}}
    \makebox[\linewidth][c]{\small\textit{IXITiny}} 
  \end{minipage}%
  \begin{minipage}{.3\linewidth}
    \centering
    \raisebox{-\height}{\includegraphics[width=\linewidth]{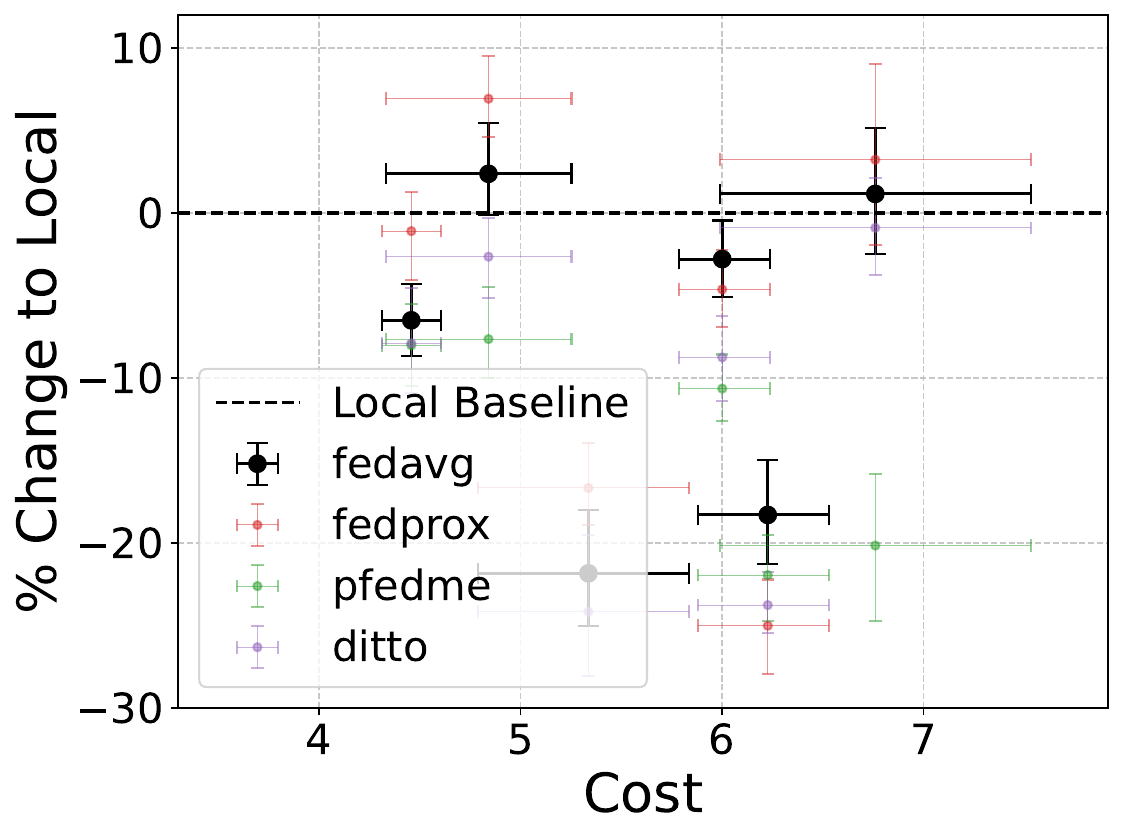}}
    \makebox[\linewidth][c]{\small\textit{ISIC2019}} 
  \end{minipage}
}

\caption{\textbf{Performance across varying Wasserstein distances}. Percentage improvement over local training baseline, with FedAvg (black) as the primary comparison. FedProx (red), pFedME (green), and Ditto (purple) are shown with reduced opacity.}
\label{fig:wass_performance}

\end{figure*}
Our metric offers three key advantages for FL compared to the original Wasserstein distance also computed on final round activations: enhanced interpretability and greater relevance to training dynamics. Firstly, unlike original Wasserstein distance, our metric provides a consistent interpretation of similarity across datasets of varying dimension and scale. To illustrate the issue of inconsistent interpretation, consider Figure \ref{fig:wass_performance} which shows results using the original Wasserstein distance. While both metrics show a connection between increasing costs and performance, the key difference is interpretability. For example, with Wasserstein distance, a cost of $\sim 5$ can imply improved learning in one dataset (ISIC) but negative learning in another (IXITiny), whereas our metric is bounded between [0,1] across domains. Second, our metric is more tightly coupled with weight divergence, a key factor influencing FL performance. 

\vspace{2mm}
\textbf{Discussion and Limitations} Our approach has several limitations that also point toward promising directions for future work. First, our metric assumes concordant feature spaces. While this holds for most FL tasks, real-world settings often involve multi-modal or partially overlapping features. Extending our framework to handle such cases, for example through Gromov–Wasserstein or related methods that relax feature correspondence, would improve its applicability. Second, our reliance on early-round representations assumes that models quickly learn informative features; when this does not hold, similarity estimates may be less reliable, though this can be mitigated by continued training until loss stabilizes. Third, we implicitly assume data stationarity across rounds, yet evolving distributions could alter similarity scores over time. In addition, class imbalance poses a practical concern as it can distort transport costs and bias similarity estimates and subsequent collaboration decisions. Finally, our evaluation focused on a single strong baseline; broader comparison to additional personalized FL methods would provide clearer context. More generally, our framework depends on several heuristic design choices (\emph{e.g.,} cost thresholds) that, while effective in our experiments, may not be universally optimal. Exploring adaptive heuristics and alternative formulations is therefore an important direction for future research.

\FloatBarrier
\clearpage 
\bibliographystyle{IEEEtran}
\bibliography{IEEEabrv,bibliography}

\end{document}